\documentclass{article}
\usepackage[utf8]{inputenc}

\PassOptionsToPackage{numbers,sort&compress}{natbib}
\usepackage[final]{neurips_2020}

\usepackage[utf8]{inputenc} 
\usepackage[T1]{fontenc}    
\usepackage{hyperref}       
\usepackage{url}            
\usepackage{booktabs}       
\usepackage{amsfonts}       
\usepackage{nicefrac}       
\usepackage{microtype}      
\usepackage{graphicx}  
\usepackage{amsmath} 
\usepackage{caption} 
\usepackage{todonotes}
\usepackage{setspace}
\usepackage{bbm}

\newcommand{\ovis}{\operatorname{OVIS}}
\newcommand{\ovismc}{\ovis_{\operatorname{MC} } }
\newcommand{\ovisinf}{\ovis_{
\boldsymbol{\sim}}}

\newcommand{\defeq}{\mathrel{\mathop:}=}
\newcommand{\Ex}{\mathbb{E}}
\newcommand{\pth}{p_\theta}
\newcommand{\qphi}{q_\phi}
\newcommand{\zb}{\mathbf{z}}
\newcommand{\eb}{\boldsymbol{\epsilon}}
\newcommand{\xb}{\mathbf{x}}
\newcommand{\vk}{v_k}
\newcommand{\znok}{\zb_{-k}}

\newcommand{\fk}{f_k}
\newcommand{\fnok}{f_{-k}}
\newcommand{\fl}{f_{l}}
\newcommand{\fnol}{f_{-l}}

\newcommand{\hk}{\mathbf{h}_k}
\newcommand{\normhk}{\| \hk \|}
\newcommand{\hl}{\mathbf{h}_{l}}
\newcommand{\hone}{\mathbf{h}_{1}}

\newcommand{\grad}{\mathbf{g}}
\newcommand{\gk}{d_k}
\newcommand{\gi}{g_i}
\newcommand{\ck}{c_k}
\newcommand{\cl}{c_l}
\newcommand{\wk}{w_{k}}
\newcommand{\wl}{w_{l}}

\newcommand{\wone}{w_{1}}
\newcommand{\wnok}{\hat{w}_{[-k]}}

\newcommand{\gphi}{\nabla_\phi}
\newcommand{\zs}{\zb_{1:K}}
 
\newcommand{\qphizs}{\qphi(\zs| \xb)} 

\newcommand{\qphizk}{\qphi(\zb_k | \xb)}
\newcommand{\qphizX}[1]{\qphi(\zb_{#1} | \xb)}
\newcommand{\iwf}{\frac{1}{K} \sum_{k=1}^K \wk}
\newcommand{\ZK}{\hat{Z}} 
\newcommand{\ZKnok}{\widetilde{Z}_{[-k]} }
\newcommand{\ZKnol}{\widetilde{Z}_{[-l]}  }
\newcommand{\ZKnokunb}{\hat{Z}_{[-k]} } 

\newcommand{\LK}{\mathcal{L}_K}
\newcommand{\order}{\mathcal{O}}
\newcommand{\expiwf}{\frac{1}{K} \sum_{k=1}^K \wk}

\newcommand{\upi}{\tilde{\pi}}
\newcommand{\upib}{\tilde{\pi}_{\beta}}
\newcommand{\pib}{\pi_{\beta}}

\newcommand{\Var}{\operatorname{Var}}
\newcommand{\Cov}{\operatorname{Cov}}
\newcommand{\softmax}{\operatorname{softmax}}
\newcommand{\ess}{\operatorname{ESS}}

\newcommand{\ws}{\operatorname{ws}}
\newcommand{\rws}{\operatorname{RWS}}
\newcommand{\tvo}{\operatorname{TVO}}
\newcommand{\vimco}{\operatorname{VIMCO}}
\newcommand{\iwae}{\operatorname{IWAE}}

\newcommand{\elbo}{\operatorname{ELBO}}
\newcommand{\snr}{\operatorname{SNR}}
\newcommand{\dsnr}{\operatorname{DSNR}}
\newcommand{\kl}{\mathcal{D}_{\mathrm{KL}}}

\DeclareMathOperator*{\argmax}{argmax}

\title{Optimal Variance Control of the Score Function Gradient Estimator for Importance Weighted Bounds}

\author{Valentin Li\'evin \textsuperscript{1} ~~~~~~~~~ Andrea Dittadi\textsuperscript{1} ~~~~~~~~~
Anders Christensen\textsuperscript{1} ~~~~~~~~~
  \textbf{Ole Winther\textsuperscript{1, 2, 3}} \\
\textsuperscript{1} Section for Cognitive Systems, Technical University of Denmark\\ \textsuperscript{2} Bioinformatics Centre, Department of Biology, University of Copenhagen \\
\textsuperscript{3} Centre for Genomic Medicine, Rigshospitalet, Copenhagen University Hospital \\
\texttt{
\{valv,adit\}@dtu.dk,~anders.christensen321@gmail.com,~olwi@dtu.dk} 
}

\begin{document}

\maketitle


\begin{abstract}
     This paper introduces novel results for the score function gradient estimator of the importance weighted variational bound (IWAE). We prove that in the limit of large $K$ (number of importance samples) one can choose the control variate such that the Signal-to-Noise ratio (SNR) of the estimator grows as $\sqrt{K}$. This is in contrast to the standard pathwise gradient estimator where the SNR decreases as $1/\sqrt{K}$. Based on our theoretical findings we develop a novel control variate that extends on VIMCO. Empirically, for the training of both continuous and discrete generative models, the proposed method yields superior variance reduction, resulting in an SNR for IWAE that increases with $K$ without relying on the reparameterization trick. The novel estimator is competitive with state-of-the-art reparameterization-free gradient estimators such as Reweighted Wake-Sleep (RWS) and the thermodynamic variational objective (TVO) when training generative models.
\end{abstract}

\section{Introduction}

Gradient-based learning is now widespread in the field of machine learning, in which recent advances have mostly relied on the backpropagation algorithm, the workhorse of modern deep learning. In many instances, for example in the context of unsupervised learning, it is desirable to make models more expressive by introducing stochastic latent variables. Backpropagation thus has to be augmented with methodologies for marginalization over latent variables.

Variational inference using an inference model (amortized inference) has emerged as a key method for training and inference in latent variable models~\citep{Blei2003-gl, Kingma2013-ey, Rezende2014-va, Bowman2015-jt, Van_den_Oord2017-sg, Higgins2017-tj, Maalo_e2019-eo}. The pathwise gradient estimator, based on the reparameterization trick~\citep{Kingma2013-ey, Rezende2014-va}, often gives low-variance estimates of the gradient for continuous distributions. 
However, since discrete distributions cannot be reparameterized, these methods are not applicable to inference in complex simulators with discrete variables, such as reinforcement learning or advanced generative processes~\cite{Sutton2000-gp, Le2018-kj, Eslami2016-wo, Miao2016-eq}.
While the score function (or Reinforce) estimator~\cite{williams1992simple} is more generally applicable, it is well known to suffer from large variance. Consequently, most of the recent developments focus on reducing the variance using control variates~\citep{Mnih2014-ev, Mnih2016-vr, Tucker2017-vy, Maddison2016-rl, gregor2013deep, gu2015muprop} and using alternative variational objectives~\citep{Hinton1995-qk, Bornschein2014-my, Le2018-kj, Masrani2019-nq}. 

Recently, variational objectives tighter than the traditional evidence lower bound (ELBO) have been proposed~\cite{Burda2015-wt,Masrani2019-nq}. In importance weighted autoencoders (IWAE)~\cite{Burda2015-wt} the tighter bound comes with the price of a $K$-fold increase in the required number of samples from the inference network. Despite yielding a tighter bound, using more samples can be detrimental to the learning of the inference model~\citep{Rainforth2018-pw}. In fact, the Signal-to-Noise ratio (the ratio of the expected gradient to its standard deviation) of the pathwise estimator has been shown to decrease at a rate $\order(K^{-1/2})$~\citep{Rainforth2018-pw}. Although this can be improved to $\order(K^{1/2})$ by exploiting properties of the gradient to cancel high-variance terms~\citep{Tucker2018-nr}, the variational distributions are still required to be reparameterizable. 
In this work we introduce $\ovis$ (\emph{Optimal Variance -- Importance Sampling}), a novel score function-based estimator for importance weighted objectives with improved $\snr$.

The main contributions of this paper are: 1) A proof that, with an appropriate choice of control variate, the score function estimator for the IWAE objective can achieve a Signal-to-Noise Ratio $\snr = \mathcal{O}(K^{1/2})$ as the number of importance samples $K\to \infty$. 2) A derivation of $\ovis$, a class of practical low-variance score function estimators following the principles of our theoretical analysis. 3) State-of-the-art results on a number of non-trivial benchmarks for both discrete and continuous stochastic variables, with comparison to a range of recently proposed score function methods.

\section{Optimizing the Importance Weighted Bound}

\paragraph{Importance weighted bound (IWAE)}

Amortized variational inference allows fitting a latent variable model $\pth(\xb, \zb)$ to the data using an approximate posterior $\qphi(\zb | \xb)$~\citep{Kingma2013-ey}. 
By using multiple importance weighted samples, we can derive a lower bound to the log marginal likelihood that is uniformly tighter as the number of samples, $K$, increases~\citep{Burda2015-wt}.
The \emph{importance weighted bound} ($\iwae$) for one data point $\xb$ is:
\begin{align}
    \LK(\xb) \defeq \Ex \left[ \log \ZK \right] \qquad \ZK \defeq \frac{1}{K} \sum_{k=1}^K \wk \qquad \wk \defeq \frac{\pth(\xb, \zb_k)}{\qphi(\zb_k | \xb)} \ ,
    \label{eq:iwae_bound}
\end{align}
where $\Ex$ denotes an expectation over the $K$-copy variational posterior $\qphizs \defeq \prod_{k=1}^{K} \qphizk$. This bound coincides with the traditional evidence lower bound (ELBO) for $K=1$. 
The log likelihood lower bound for the entire data set is $\LK(\xb_{1:n})=\sum_{i=1}^n \LK(\xb_i)$. In the following we will derive results for one term $\LK = \LK(\xb)$.

\paragraph{Score function estimator}

Without making assumptions about the variational distribution, the gradient of the importance weighted bound~\eqref{eq:iwae_bound} with respect to the parameters of the approximate posterior factorizes as (see Appendix~\ref{apdx:grads}):
\begin{equation}
    \gphi \LK = \Ex
    \left[\sum\nolimits_k  \gk \hk \right] 
    \qquad
    \gk \defeq \log \ZK - \vk
    \qquad
    \vk \defeq \frac{\wk}{\sum_{l=1}^K \wl} \ ,
    \label{eq:gradients}
\end{equation}
where $\hk \defeq \gphi \log \qphizk$ is the score function. 
A Monte Carlo estimate of the expectation in~\eqref{eq:gradients} yields the \emph{score function} (or \emph{Reinforce}) \emph{estimator}.

\paragraph{Control variates}

The vanilla score function estimator of~\eqref{eq:gradients} is often not useful in practice due to its large sample-to-sample variance. By introducing control variates that aim to cancel out zero expectation terms, this variance can be reduced while keeping the estimator unbiased.

Given posterior samples $\zb_1, \ldots, \zb_K \sim \qphizs$, let $\zb_{-k}$ denote $[\zb_1, \ldots, \zb_{k-1}, \zb_{k+1}, \ldots, \zb_K]$, let $\Ex_k[\ldots]$ and $\Ex_{-k}[\ldots]$ be the expectations over the variational distributions of $\zb_k$ and $\znok$, respectively, and let $\{ \ck \}_{k=1}^K$ be scalar control variates, with each $\ck=\ck(\zb_{-k})$ independent of $\zb_k$. 
Using the independence of $\ck$ and $\hk$ for each $k$, and the fact that the score function has zero expectation, we have $\Ex[\ck \hk] = \Ex_{-k}[\ck] \Ex_k[\hk] = 0$. Thus, we can define an unbiased estimator of~\eqref{eq:gradients} as:%
\begin{gather}
    \grad  \defeq \sum\nolimits_k
    \left( \gk -\ck \right) \hk 
    \label{eq:estimator_definition} \\
    \Ex[\grad] 
    = \Ex \left[\sum\nolimits_k
        \left( \gk - \ck \right) \hk \right] 
    = \Ex \left[\sum\nolimits_k
        \gk  \hk \right] 
    = \gphi \LK \ .
\end{gather}

In the remainder of this paper, we will use the decomposition $\gk = \fk + \fnok$, where $\fk=\fk(\zb_k,\zb_{-k})$ and $\fnok=\fnok(\zb_{-k})$ denote terms that depend and do not depend on $\zb_k$, respectively.
This will allow us to exploit the mutual independence of $\{\zb_k\}_{k=1}^K$ to derive optimal control variates.

\paragraph{Signal-to-Noise Ratio (SNR)}

We will compare the different estimators on the basis of their Signal-to-noise ratio. Following~\citep{Rainforth2018-pw}, we define the $\snr$ for each component of the gradient vector as%
\begin{equation}
\snr_i \defeq \frac{|\Ex[ \gi]|}{\sqrt{\Var[\gi]}} \ ,    
\end{equation}
where $\gi$ denotes the $i$th component of the gradient vector.

In Section~\ref{sec:asymptotic} we derive the theoretical $\snr$ for the optimal choice of control variates in the limit $K \to \infty$. In Section~\ref{sec:optimal_control} we derive the optimal scalar control variates $\{\ck\}_{k=1}^K$ by optimizing the trace of the covariance of the gradient estimator $\grad$,
and in Section~\ref{sec:results} we experimentally compare our approach with state-of-the-art gradient estimators in terms of $\snr$.

\section{Asymptotic Analysis of the Signal-to-Noise Ratio} \label{sec:asymptotic}

Assuming the importance weights have finite variance, i.e. $\Var[\wk]< \infty$, we can derive the asymptotic behavior of the $\snr$ as $K \to \infty$ by expanding $\log \ZK$ as a Taylor series around $Z\defeq \pth(\xb) = \int \pth(\xb, \zb) d\zb$~\cite{Rainforth2018-pw}.
A direct application of the pathwise gradient estimator (reparameterization trick) to the importance weighted bound results in an $\snr$ that scales as $\order(K^{-1/2})$~\citep{Rainforth2018-pw}, which can be improved to $\order(K^{1/2})$ by exploiting properties of the gradient~\citep{Tucker2018-nr}.
In the following we will show that, for a specific choice of control variate,
the $\snr$ of the score function estimator scales as $\order(K^{1/2})$. 
Thus, a score function estimator exists for which \emph{increasing the number of importance samples 
benefits the gradient estimate of the parameters of the variational distribution}.

For the asymptotic analysis we rewrite the estimator as 
$\grad = \sum_k \big( -  \frac{\partial \log \ZK}{\partial w_k} w_k + \log \ZK - \ck \big) \hk$ 
and apply a second-order Taylor expansion to $\log \ZK$.
The resulting expression $\grad = \sum_k (f_k + f_{-k} - c_k)\hk$ separates terms $f_k$ that contribute to the expected gradient from terms $f_{-k}$ that have zero expectation and thus only contribute to the variance (cf. Appendix~\ref{app:asymptotic_analysis}):
\begin{gather}
    f_k \approx \frac{\wk^2}{2 K^2 Z^2} \\ 
    f_{-k} \approx \log Z - \frac{3}{2} + \frac{2}{KZ} \sum\nolimits_{l \neq k} \wl - \frac{1}{2 K^2 Z^2} \left( \sum\nolimits_{l \neq k} \wl \right)^2 \ .
\end{gather}%
Since $\fnok$ and $\ck$ are independent of $\hk$, the expected gradient is (cf. Appendix~\ref{apds:asymptoticgrad:expectation}):
\begin{align}
    \Ex[\grad] = \sum\nolimits_k \Ex[f_k \hk] \approx \frac{1}{2Z^2K} \Ex_1 \left[ \wone^2 \hone \right] = \order(K^{-1})\ , 
\end{align}
where $\Ex_1$ denotes an expectation over the first latent distribution $\qphizX{1}$.
Since the choice of control variates $c_k=c_k(\zb_{-k})$ is free, we can choose $c_k=f_{-k}$ to cancel out all zero expectation terms. The resulting covariance, derived in Appendix~\ref{apds:asymptoticgrad:variance}, is:
\begin{align}
\Cov[\grad] = \Cov \left[\sum\nolimits_k f_k \hk \right] \approx
\frac{1}{4 K^3 Z^4 } \Cov_1 \left[\wone^2 \hone \right] = \order(K^{-3})
\end{align}
with $\Cov_1$ indicating the covariance over $\qphizX{1}$.
Although as we discuss in Section~\ref{sec:optimal_control} this is not the minimal variance choice of control variates, it is sufficient to achieve an $\snr$ of $\order(K^{1/2})$.

\section{Optimal Control Variate} \label{sec:optimal_control}

The analysis above shows that in theory it is possible to attain a good SNR with the score function estimator. In this section we derive the optimal (in terms of variance of the resulting estimator) control variates $\{c_k\}_{k=1}^K$ by decomposing $\grad=\sum_k (f_k + f_{-k} - c_k)\hk$ as above, and minimizing the trace of the covariance matrix, i.e. $\Ex[||\grad||^2] - ||\Ex [\grad]||^2$.
Since $\Ex[\fnok \hk]$ and $\Ex[\ck \hk]$ are both zero, $\Ex [\grad] = \gphi \LK$ does not depend on $\ck$. Thus, the minimization only involves the first term:
\begin{align*}
  \frac{1}{2}\frac{\partial }{\partial \ck} \Ex\left[||\grad||^2\right] 
  &= \Ex\left[ \hk^T \sum\nolimits_l (\fl + \fnol - \cl ) \hl \right] \\ 
  &= \Ex_{-k}\left[ \sum\nolimits_l \Ex_k\left[ \fl \hk^T \hl \right] + ( \fnok - \ck ) \Ex_{k}\left[ \normhk^2 \right] \right] \ .
\end{align*}
where $\Ex_k$ and $\Ex_{-k}$ indicate expectations over $\qphizk$ and $\qphizX{-k}$, respectively. Setting the argument of $\Ex_{-k}$ to zero, we get the optimal control variates $\ck=\ck(\zb_{-k})$ and gradient estimator $\grad$:%
\begin{align}
    \ck & = \fnok + \sum\nolimits_l \frac{\Ex_k \left[ \fl \hk^T \hl \right]}{\Ex_k \left[ \normhk^2   \right]} \label{eq:copt} \\
    \grad & = \sum\nolimits_k \left(\fk - \sum\nolimits_l \frac{\Ex_k \left[ \fl \hk^T \hl \right]}{\Ex_k \left[ \normhk^2   \right]} \right) \hk
    \ .   \label{eq:gradopt}
\end{align}
Applying~\eqref{eq:gradopt} in practice requires marginalizing over one latent variable and decoupling terms that do not depend on $\zb_k$ from those that do. In the remainder of this section we will 1) make a series of approximations to keep computation tractable, and 2) consider two limiting cases for the \emph{effective sample size} (ESS)~\citep{kong1992note} in which we can decouple terms.

\paragraph{Simplifying approximations to Equation~(\ref{eq:gradopt})}

First, we consider a term with $l\neq k$, define $\Delta \fl \defeq \fl- \Ex_k[f_l]$, and subtract and add $\Ex_k[f_l]$ from inside the expectation:
\begin{align*}
    \Ex_k \left[ \fl \hk^T\right]  \hl  = \Ex_k \left[ \Delta \fl  \hk^T \right] \hl + \Ex_k[f_l] \Ex_k  \left[ \hk^T \right] \hl = \Ex_k \left[ \Delta \fl \hk^T \right] \hl
\end{align*}
where we used the fact that $\Ex_k  \left[ \hk \right] = 0$.
The $l\neq k$ terms thus only contribute to fluctuations relative to a mean value, and we assume they can be neglected.

Second, we assume that $|\phi|$, the number of parameters of $q_{\phi}$, is large, and the terms of the sum $\normhk^2 = \sum_{i=1}^{|\phi|} h_{ki}^2$ are approximately independent with finite variances $\sigma_i^2$. By the Central Limit Theorem 
we approximate the distribution of $\Delta \normhk^2 \defeq \normhk^2 - \Ex_k \left[ \normhk^2 \right]$ with a zero-mean Gaussian with standard deviation $\big(\sum_{i=1}^{|\phi|} \sigma_i^2)^{1/2}$.
Seeing that $\Ex_k \left[ \normhk^2 \right]$ is $\order(|\phi|)$, we have
\begin{equation*}
    \frac{\Ex_k \left[ \fk \normhk^2 \right]}{\Ex_k \left[ \normhk^2   \right]} = \Ex_k \left[ \fk \right] + \frac{\Ex_k \left[ \fk \Delta \normhk^2 \right]}{\Ex_k \left[ \normhk^2   \right]}  = \Ex_k \left[ \fk \right] + \mathcal{O}(|\phi|^{-1/2})  \ ,
\end{equation*}
where we used that the argument in the numerator scales as $\big( \sum_{i=1}^{|\phi|} \sigma_i^2 \big)^{1/2} =\order(|\phi|^{1/2})$.

Finally, the expectation can be approximated with a sample average. Writing $f_k = f_k(\zb_k,\zb_{-k})$ and drawing $S$ new samples $\zb^{(1)},\dots, \zb^{(S)} \sim \qphi(\zb | \xb)$:
\begin{equation*}
    \Ex_k \left[ \fk \right] \approx \frac{1}{S} \sum_{s=1}^S f_k(\zb^{(s)},\zb_{-k}) \ .    
\end{equation*}
This will introduce additional fluctuations with scale $S^{-1/2}$.

Putting these three approximations together and using $\gk(\zb_k,\zb_{-k}) = f_k(\zb_k,\zb_{-k}) + f_{-k}(\zb_{-k})$, we obtain the sample-based expression of the $\ovis$ estimator, called $\ovismc$ in the following:
\begin{equation}
     \ovismc : \quad \grad \approx \sum_k \left(\gk(\zb_k,\zb_{-k})  - \frac{1}{S} \sum_{s=1}^S \gk(\zb^{(s)},\zb_{-k}) \right) \hk
    \ . 
    \phantom{\ovismc : \quad }
    \label{eq:sample-based-control} 
\end{equation}

 Naively, this will produce a large computational overhead because we now have in total $KS$ terms. However, we can reduce this to $\mathcal{O}(K+S)$ because the bulk of the computation comes from evaluating the importance weights and because the $S$ auxiliary samples can be reused for all $K$ terms.

\paragraph{Effective sample size (ESS)}

The ESS~\citep{kong1992note} is a commonly used yardstick of the efficiency of an importance sampling estimate, defined as
\begin{equation}
    \ess \defeq \frac{\left( \sum_k \wk \right)^2}{\sum_k \wk^2} = \frac{1}{\sum_k v_k^2} \in [1,K] \ .
\end{equation}
A low ESS occurs when only a few weights dominate, which indicates that the proposal distribution $q$ poorly matches $p$.
In the opposite limit, the variance of importance weights is finite and the ESS will scale with $K$.
Therefore the limit $\ess\gg 1$ corresponds to the asymptotic limit studied in Section~\ref{sec:asymptotic}. 

\paragraph{Optimal control for ESS limits and unified interpolation}

In the following, we consider the two extreme limits $\ess\gg 1$ and $\ess \approx 1$ to derive sample-free approximations to the optimal control. We can thus in these limits avoid the sample fluctuations and excess computation of $\ovismc$.

We first consider $\ess\gg 1$ and for each $k$ we introduce the unnormalized leave-$\wk$-out approximation to $\ZK$:%
\begin{equation}
\ZKnok \defeq \frac{1}{K}\sum_{l\neq k} \wl \quad\text{such that}\quad \ZK - \ZKnok = \frac{\wk}{K} \ .
\end{equation}%
Assuming $\Var [\wk] < \infty$, this difference is $\order(K^{-1})$ as $K \to \infty$, thus we can expand $\log \ZK$ around $\ZK = \ZKnok$. In this limit, the optimal control variate simplifies to  (cf.~Appendix~\ref{apds:ESSlarge}):%
\begin{equation}
    \ess \gg 1 : \quad\ck \approx \log \frac{1}{K-1} \sum_{l\neq k} \wl + \log(1 - \frac{1}{K})\ .
    \phantom{\ess \gg 1 : \quad}
    \label{eq:control-variate-ess>>1}
\end{equation}
When $\ess \approx 1$, one weight is much larger than the others and the assumption above is no longer valid. To analyze this frequently occurring scenario, assume that $k' = \argmax_{l} \wl$ and $w_{k'} \gg \sum_{l \neq k'} \wl$. 
In this limit $\log \ZK \approx \log w_{k'}/K$ and $v_k \approx \delta_{k,k'}$ and thus  $d_{k} = \log w_{k'}/K - \delta_{k,k'}$. 
In Appendix~\ref{apds:ESS1} we show we can approximate Equation (\ref{eq:copt}) with
\begin{equation}
    \ess \approx 1 : \qquad  \ck \approx \log \frac{1}{K-1} \sum_{l \neq k} \wl - v_k \ . 
    \phantom{\ess \approx 1 : \qquad }
    \label{eq:control-variate-ess=1}
\end{equation}
We introduce $\ovisinf$  to interpolate between the two limits (Appendix~\ref{apds:unified-control}):
\begin{equation}
    \quad \ck^{\gamma} \defeq \log \frac{1}{K-1} \sum_{l \neq k} \wl - \gamma  \vk + (1 - \gamma)  \log \left(1 - \frac{1}{K} \right)  \qquad \gamma \in [0, 1] \ .
    \label{eq:alpha-control}
\end{equation}
In this paper we will only conduct experiments for the two limiting cases $\gamma=0$, corresponding to Equation~\eqref{eq:control-variate-ess>>1}, and $\gamma=1$ approximating Equation~\eqref{eq:control-variate-ess=1}. Tuning the parameter $\gamma$ in the range $[0,1]$ will be left for future work. We discuss the implementation in the appendix \ref{apdx:implementation}.

\paragraph{Higher ESS with looser lower bound}\label{sec:renyi}

Empirically we observe that training may be impaired by a low $\ess$ and by \emph{posterior collapse}~\citep{Bowman2015-jt, Chen2016-gx, Sonderby2016-uv, Kingma2016-kf, Dieng2018-mf}. This motivates trading the tight IWAE objective for a gradient estimator with higher $\ess$. To that end, we use the importance weighted Rényi (IWR) bound:
\begin{equation}
    \LK^{\alpha}(\xb) \defeq \frac{1}{1 - \alpha} \Ex \left[ \log \ZK(\alpha) \right] \quad \ZK(\alpha) \defeq \frac{1}{K} \sum\nolimits_k \wk^{1 - \alpha} 
\end{equation}
which for $\alpha \in [0, 1]$ is a lower bound on the Rényi objective $\log \Ex_1 \left[ w_1^{1-\alpha} \right]/(1-\alpha)$~\cite{Li2016-ph}. The Rényi objective in itself coincides with $\log p(\xb)$ for $\alpha=0$ and is monotonically non-increasing in $\alpha$, i.e.~is an evidence lower bound~\cite{Li2016-ph}. So we have a looser bound but higher $\ess(\alpha)=1/\sum_k v_k^2(\alpha) \geq \ess(0)$ for $\alpha \in [0, 1]$ with $v_k(\alpha)= w_k^{1-\alpha}/\sum_l w_l^{1-\alpha}$. Furthermore, for $\alpha = 1$ the bound corresponds to the ELBO and the divergence $\kl(\qphi(\zb | \xb) || \pth(\zb|\xb))$ is guaranteed to be minimized. In Appendix~\ref{apdx:beta-bound} we derive the score function estimator  and control variate expressions for $\LK^{\alpha}$. The objective can either be used in a warm-up scheme by gradually decreasing $\alpha\to 0$ throughout iterations or can be run with a constant $0<\alpha<1$.

\section{Related Work}

The score function estimator with control variates can be used with all the commonly used variational families. By contrast, the reparameterization trick is only applicable under specific conditions. We now give a brief overview of the existing alternatives and refer the reader to~\cite{Mohamed2019-bx} for a more extensive review. The importance of handling discrete distributions without relaxations is discussed in~\cite{Le2018-kj}.

NVIL~\cite{Mnih2014-ev}, DARN~\cite{gregor2013deep}, and MuProp~\cite{gu2015muprop} demonstrate that score function estimators with carefully crafted control variates allow to train deep generative models. 
$\vimco$~\cite{Mnih2016-vr} extends this to multi-sample objectives, and recycles the Monte Carlo samples $\znok$ to define a control variate $\ck = \ck(\znok)$. Unlike $\ovis$, $\vimco$ 
only controls the variance of the term $\log \ZK$ in $\gk = \log \ZK - \vk$, leaving $\vk$ uncontrolled, and
causing the $\snr$ to decrease with the number of particles $K$ as we empirically observe in Section~\ref{exp:asymptotic-variance}.  We provide a detailed review of $\vimco$ in Appendix~\ref{apdx:estimators}.

The Reweighted Wake-Sleep (RWS) algorithm~\cite{Bornschein2014-my} is an extension of the original Wake-Sleep algorithm ($\ws$)~\cite{Hinton1995-qk} that alternates between two distinct learning phases for optimizing importance weighted objectives.
A detailed review of $\rws$ and $\ws$ is available in Appendix~\ref{apdx:estimators}.

The Thermodynamic Variational Objective ($\tvo$)~\cite{Masrani2019-nq} is a lower bound to $\log \pth(\xb)$ that stems from a Riemannian approximation of the Thermodynamic Variational Identity (TVI), and unifies the objectives of Variational Inference and Wake-Sleep. Evaluating the gradient involves differentiating through an expectation over a distribution with an intractable normalizing constant. To accommodate this, the authors propose an estimator that generalizes the score function estimator based on a tractable covariance term. 
We review the $\tvo$ in more detail in Appendix~\ref{apdx:estimators}.

Given a deterministic \textit{sampling path} $g(\eb ; \theta)$ such that $\zb \sim \pth(\zb)$ and $\zb = g(\eb ; \theta), \eb \sim p(\eb)$ are equivalent, one can derive a \emph{pathwise gradient estimator} of the form $\nabla_{\theta} \Ex_{\pth(\zb)} \left[ f_{\theta} ( \zb ) \right] = \Ex_{p(\eb)}\left[ \nabla_{\theta} f_{\theta} ( g(\eb ; \theta) ) \right]$. This  estimator -- introduced in machine learning as the \textit{reparameterization trick} or \textit{stochastic backpropagation}~\citep{Kingma2013-ey, Rezende2014-va} -- exhibits low variance thanks to the structural information provided by the sampling path.
Notably, a zero expectation term can be removed from the estimator~\citep{Roeder2017-qc}. Extending on this, \citep{Tucker2018-nr} derives an alternative gradient estimator for $\iwae$ that exhibits $\snr \sim K^{1/2}$, as opposed to $\snr \sim K^{-1/2}$ for the \emph{standard} $\iwae$ objective~\cite{Rainforth2018-pw}. 

Continuous relaxations of discrete distributions yield a biased low-variance gradient estimate thanks to the reparameterization trick~\citep{Maddison2016-rl, Jang2016-vb}.
Discrete samples can be obtained using the Straight-Through estimator~\citep{Bengio2013-ui, Van_den_Oord2017-sg}. The resulting gradient estimate remains biased, but can be used as a control variate for the score function objective, resulting in an unbiased low-variance estimate of the gradient~\citep{Tucker2017-vy, Grathwohl2017-mz}.

\section{Experimental Results}\label{sec:results}

We conduct a number of experiments\footnote{The full experimental framework is available at \href{https://github.com/vlievin/ovis}{github.com/vlievin/ovis}} on benchmarks that have previously been used to test score function based estimators.
All models are trained via stochastic gradient ascent using the Adam optimizer~\citep{Kingma2014-cn} with default parameters. We use regular gradients on the training objective for the generative model parameters $\theta$. The $\snr$ for $\theta$ scales as $\mathcal{O}(K^{1/2})$ \cite{Rainforth2018-pw}. 

\subsection{Asymptotic Variance}\label{exp:asymptotic-variance}

\begin{figure}[h]
        \centering
        \includegraphics[width=1.\linewidth]{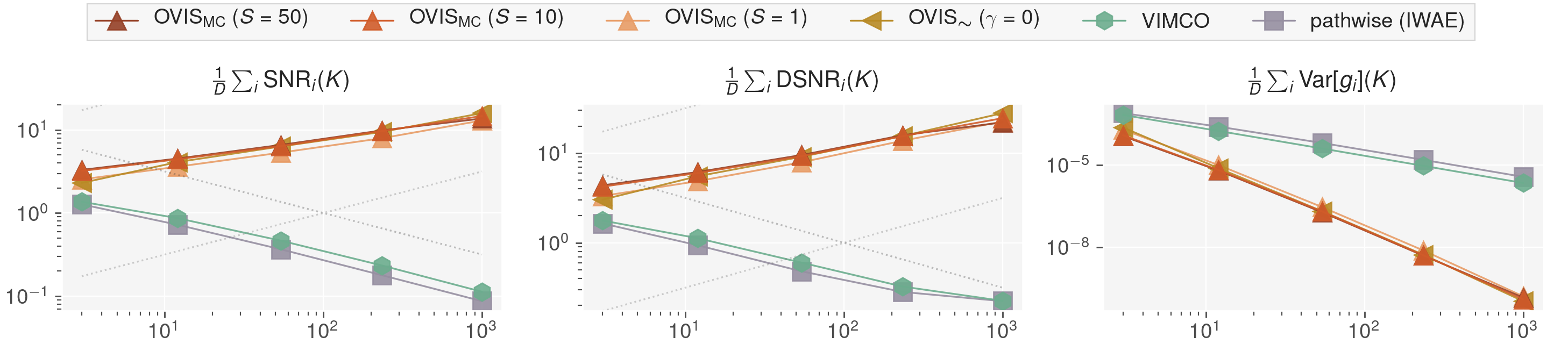}
  \caption{Gaussian model. Parameter-wise average of the asymptotic $\snr$, $\dsnr$ and variance of the gradients of the parameter $b$ for different number of particles $K \in [3, 1000]$ using $10^4$ MC samples. The dotted lines stand for $y = 10^{\pm 1} K^{\pm 0.5}$. }\label{fig:asymptotic-variance}
\end{figure}
Following~\cite{Rainforth2018-pw}, we empirically corroborate the asymptotic properties of the $\ovis$ gradient estimator by means of the following simple model: 
\begin{equation*}
    \zb \sim \mathcal{N}(\zb ; \boldsymbol{\mu}, \mathbf{I}), \quad \xb | \zb \sim \mathcal{N}(\xb ; \zb, \mathbf{I}), \quad \qphi(\zb | \xb)=\mathcal{N}\left(\zb ; \mathbf{A} \xb+ \mathbf{b}, \tfrac{2}{3} \mathbf{I}\right).
\end{equation*}
where $\xb$ and $\zb$ are real vectors of size $D=20$.
We sample $N=1024$ points $\left\{\xb^{(n)}\right\}_{n=1}^{N}$ from the \emph{true} model where $\boldsymbol{\mu}^\star \sim \mathcal{N}(\mathbf{0}, \mathbf{I})$. The optimal parameters are $\mathbf{A}^\star=\mathbf{I} / 2$, $\mathbf{b}^\star=\boldsymbol{\mu}^\star / 2$, and $\boldsymbol{\mu}^\star=\frac{1}{N} \sum_{n=1}^{N} \xb^{(n)}$. The model parameters are obtained by adding Gaussian noise of scale $\epsilon = 10^{-3}$. We measure the variance and the $\snr$ of the gradients with $10^4$ MC samples. We also measured the \emph{directional} $\snr$ ($\dsnr$~\citep{Rainforth2018-pw}) to probe if our results hold in the multidimensional case.

In Figure~\ref{fig:asymptotic-variance} we report the gradient statistics for $\mathbf{b}$. We observe that using more samples in the standard $\iwae$ leads to a decrease in $\snr$ as $\order(K^{-1/2})$ for both $\vimco$ and the pathwise-$\iwae$ \cite{Rainforth2018-pw}. The tighter variance control provided by $\ovis$ leads the variance to decrease almost at a rate $\order(K^{-3})$, resulting in a measured $\snr$ not far from $\order(K^{1/2})$ both for $\ovismc$ and $\ovisinf$. This shows that, despite the approximations, the proposed gradient estimators $\ovismc$ and $\ovisinf$ are capable of achieving the theoretical $\snr$ of $\order(K^{1/2})$ derived in the asymptotic analysis in Section~\ref{sec:asymptotic}.

In Appendix~\ref{apdx:fit-gaussian}, we learn the parameters of the Gaussian model using $\ovis, \rws, \vimco$ and the $\tvo$. We find that optimal variance reduction translates into a more accurate estimation of the optimal parameters of the inference network when compared to $\rws, \vimco$ and the $\tvo$. 

\subsection{Gaussian Mixture Model}\label{exp:gmm}

\begin{figure}
  \centering
  \includegraphics[width=\textwidth]{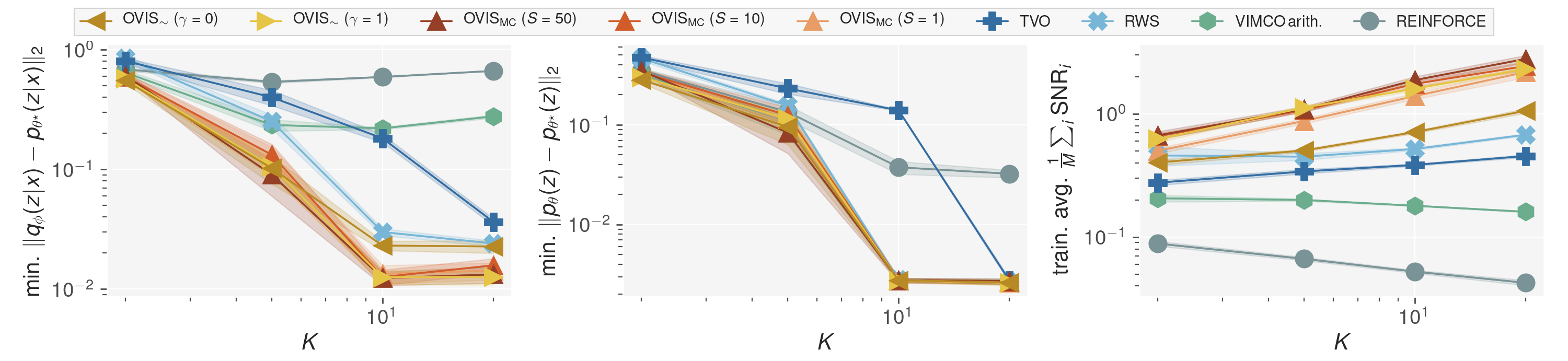}
  \caption{Training of the Gaussian mixture model. Minimum test-diagnostics recorded during training and training average of the $\snr$ of the gradients of $\phi$ with $M = \operatorname{card}(\phi)$. In contrast to $\vimco$, $\ovisinf$ and $\ovismc$ all benefit from the increase of the particles budget, $\ovismc$ yields the most accurate posterior among the compared methods. }\label{fig:gmm-pivot}
\end{figure}

We evaluate $\ovis$ on a Gaussian Mixture Model and show that, unlike $\vimco$~\citep{Le2018-kj}, our method yields better inference networks as the number of particles $K$ increases.
Following~\citep{Le2018-kj}, we define:
\begin{equation*}
p_{\theta}(z) =\operatorname{Cat}(z | \softmax(\theta)) \quad
p(x | z) = \mathcal{N}\left(x | \mu_{z}, \sigma_{z}^{2}\right) \quad
q_{\phi}(z | x) =\operatorname{Cat}\left(z | \softmax\left(\eta_{\phi}(x)\right)\right)
\end{equation*}
where $z \in \{0,\dots,C - 1\}$, $\mu_z = 10 z$, $\sigma_z =5$, and $C=20$ is the number of clusters. The inference network $\eta_{\phi}$ is parameterized by a multilayer perceptron with architecture $1$--$16$--$C$ and $\tanh$ activations. The true generative model is set to $p_{\theta^\star}(z = c) = (c+5) / \sum\nolimits_{i=1}^{C}(i+5)$.

All models are trained for $100$k steps with $5$ random seeds. We compare $\ovis$ with $\vimco$, $\rws$ with wake-$\phi$ update, Reinforce, and the $\tvo$. For the latter we chose to use 5 partitions and $\beta_{1} = 10^{-2}$, after a hyperparameter search over $\beta_1 \in \{ 10^{-1}, 10^{-1.5}, 10^{-2}, 10^{-2.5}, 10^{-3}\}$ and $\{2, 5\}$ partitions.

Each model is evaluated on a held-out test set of size $M=100$. We measure the accuracy of the learned posterior $\qphi(z | x)$ by its average $L_2$ distance from the true posterior, i.e. $\frac{1}{M} \sum\nolimits_{m=1}^{M}\left\|q_{\phi}\left(z | x^{(m)}\right)-p_{\theta^{\star}}\left(z | x^{(m)}\right)\right\|_2$. As a sanity check, we assess the quality of the generative model using $\left\|\softmax(\theta)-\softmax\left(\theta^\star \right)\right\|_2$. The $\snr$ of the gradients for the parameters $\phi$ is evaluated on one mini-batch of data using $500$ MC samples.

We report our main results in Figure~\ref{fig:gmm-pivot}, and training curves in Appendix~\ref{apdx:gmm}. In contrast to $\vimco$, the accuracy of the posteriors learned using $\ovismc$ and $\ovisinf$ all improve monotonically with $K$ and outperform the baseline estimators, independently of the choice of the number of auxiliary particles $S$. All $\ovis$ methods outperform the state-of-the-art estimators $\rws$ and the $\tvo$, as measured by the $L_2$ distance between the approximate and the true posterior. 

\subsection{Deep Generative Models}\label{exp:binary-images}

We utilize the $\ovis$ estimators to learn the parameters of both discrete and continuous deep generative models using stochastic gradient ascent. The base learning rate is fixed to $3 \cdot 10^{-4}$, we use mini-batches of size $24$ and train all models for $4 \cdot 10^6$ steps. We use the statically binarized MNIST dataset~\cite{Salakhutdinov2008-re} with the original training/validation/test splits of size 50k/10k/10k. We follow the experimental protocol as detailed in~\cite{Masrani2019-nq}, including the $\beta$ partition for the $\tvo$ and the exact architecture of the models. We use a three-layer Sigmoid Belief Network~\citep{Neal1992-fm} as an archetype of discrete generative model~\cite{Mnih2014-ev, Mnih2016-vr, Masrani2019-nq} and a Gaussian Variational Autoencoder~\cite{Kingma2013-ey} with 200 latent variables. All models are trained with three initial random seeds and for $K \in \{5, 10, 50\}$ particles. 

We assess the performance based on the marginal log-likelihood estimate $\log \hat{p}_{\theta} (\xb) = \mathcal{L}_{5000}(\xb)$, that we evaluate on $10$k \emph{training} data points, such as to disentangle the training dynamics from the regularisation effect that is specific to each method. We measure the quality of the inference network solution using the divergence $\kl \left( \qphi(\zb | \xb) || \pth(\zb | \xb) \right) \approx \log \hat{p}_{\theta}(\xb) - \mathcal{L}_1(\xb)$.  The full training curves -- including the test log likelihood and divergences -- are available in Appendix~\ref{apdx:dgm}. We will show that $\ovis$ improves over $\vimco$, on which it extends, and we show that combining $\ovisinf$ with the Variational Rényi bound (IWR) as described in Section~\ref{sec:renyi} outperforms the $\tvo$.


\subsubsection{Sigmoid Belief Network (SBN)}\label{sec:sbm}

\begin{figure}
        \centering
        \includegraphics[width=1\linewidth]{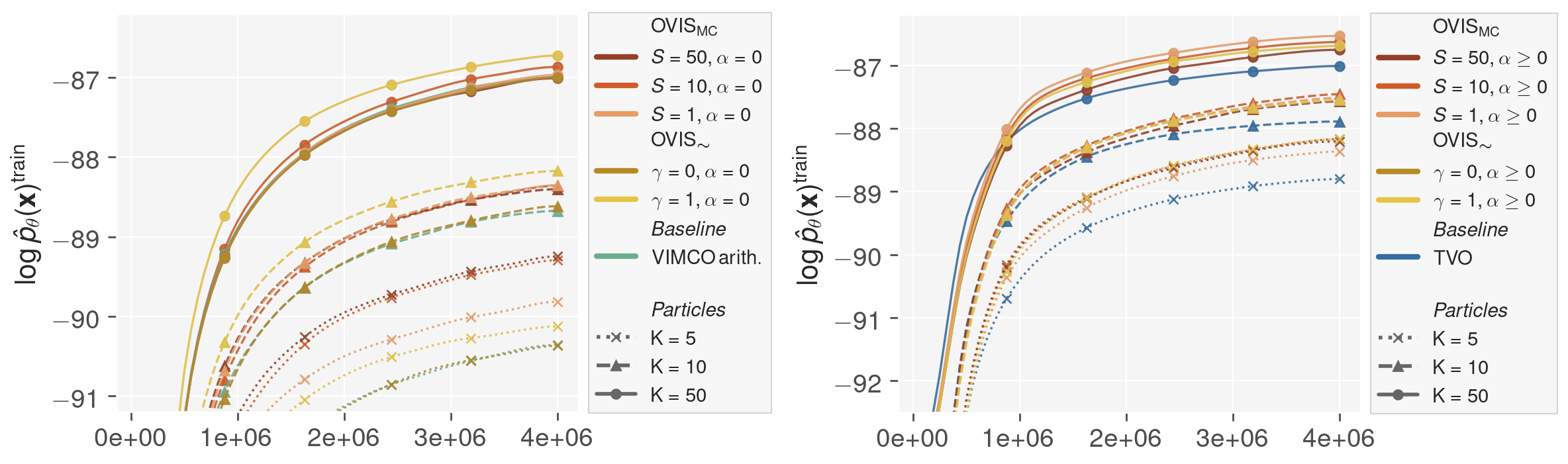}
  \caption{Training a Sigmoid Belief Network on Binarized MNIST. (Left) Optimizing for the importance weighted bound $\LK$ using $\ovis$. (Right) Optimizing for the Rényi importance lower bound $\LK^\alpha$ using $\ovis$ with $\alpha$ annealing $0.99 \to 0$. The curves are averaged over three seeds and smoothed for clarity.}\label{fig:sbm-pivot}
\end{figure}
\paragraph{A. Comparison with VIMCO}

We learn the parameters of the SBN using the $\ovis$ estimators for the IWAE bound and use $\vimco$ as a baseline. We report $\log \hat{p}_{\theta}(\xb)$ in the left plot of Figure~\ref{fig:sbm-pivot}. All $\ovis$ methods outperform $\vimco$, ergo supporting the advantage of optimal variance reduction. When using a small number of particles $K=5$, learning can be greatly improved by using an accurate MC estimate of the optimal control variate, as suggested by $\ovismc(S=50)$ which allows gaining $+1.0$ nats over $\vimco$. While $\ovis (\gamma = 0)$, designed for large $\ess$ barely improved over $\vimco$, the biased $\ovisinf(\gamma = 1)$ for low $\ess$ performed significantly better than other methods for $K \geq 10$, which coincides with the $\ess$ measured in the range $[1.0, 3.5]$ for all methods. We attribute the relative decrease of performances observed for $\ovismc$ for $K=50$ to \textit{posterior collapse}. 

\paragraph{B. Training using IWR bounds}

In Figure~\ref{fig:sbm-pivot} (right) we train the SBN using $\ovis$ and the $\tvo$. $\ovis$ is coupled with the objective $\LK^{\alpha}$ for which we anneal the parameter $\alpha$ from $0.99$ ($\LK^{0.99} \approx \mathcal{L}_1$) to $0$ ($\LK^{0} = \LK$) during $1e6$ steps using geometric interpolation. For all $K$ values, $\ovis$ outperform the $\tvo$ and $\ovisinf(\gamma=1)$ performs comparably with $\ovismc$.

\subsubsection{Gaussian Variational Autoencoder (VAE)}

\begin{figure}
        \centering
        \includegraphics[width=\linewidth]{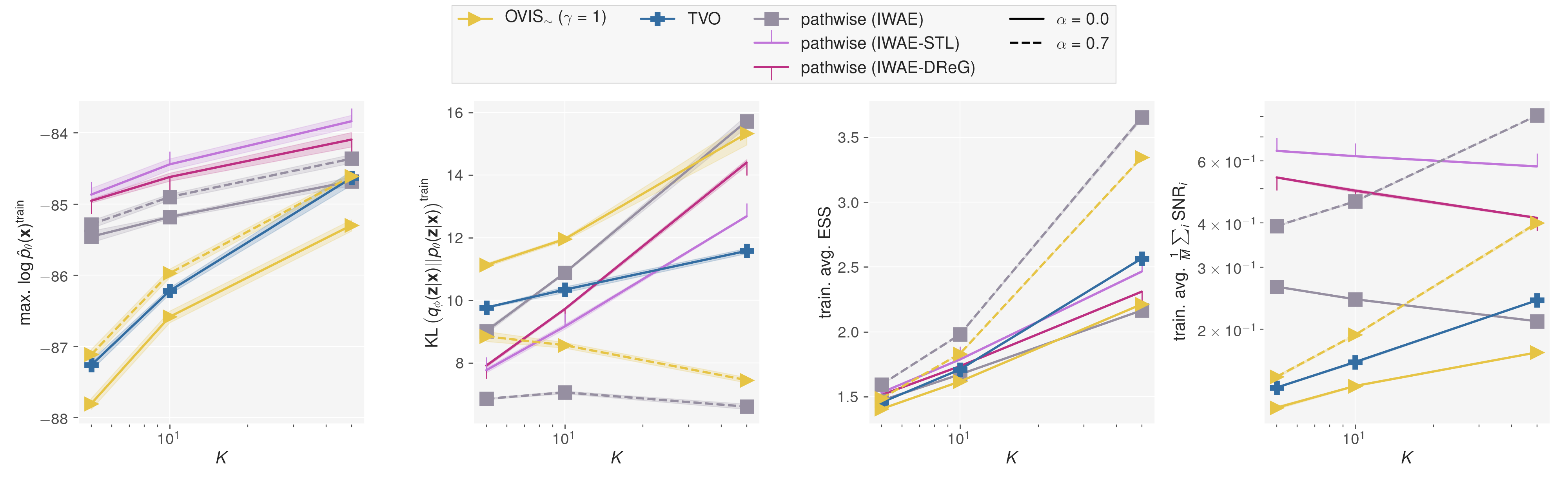}
  \caption{Training a one layer Gaussian VAE. Maximum recorded training $\log \hat{p}_{\theta}(\xb)$, final estimate of the bound $\kl \left( \qphi(\zb | \xb) || \pth(\zb | \xb) \right)$ and training average of the $\ess$ and of the $\snr$. $\ovis$ yields similar likelihood performances as the $\tvo$ but benefits from a tighter bound thanks to optimizing for the IWR bound.}\label{fig:gaussian-vae-pivot}
\end{figure}

In Figure~\ref{fig:gaussian-vae-pivot} we train the Gaussian VAE using the standard pathwise IWAE, Sticking the Landing (STL) \citep{Roeder2017-qc}, DReG \citep{Tucker2018-nr}, the $\tvo$ and $\ovisinf (\gamma=1)$. 

$\ovis$ is applied to the IWR bound with $\alpha = 0.7$. As measured by the training likelihood, $\ovisinf (\gamma=1)$ coupled with the IWR bound performs on par with the $\tvo$, which bridges the gap to the standard pathwise IWAE for $K=50$, although different objectives are at play. The advanced pathwise estimators (STL and DReG) outperform all other methods. Measuring the quality of the learned proposals $\qphi(\zb |\xb)$ using the KL divergence allows disentangling the $\tvo$ and $\ovisinf$ methods, as $\ovis(\gamma = 1)$ applied to the IWR bound outputs higher-quality approximate posteriors for all considered number of particles.

\subsection{A final Note on $\ovisinf (\gamma=1)$}

$\ovisinf (\gamma = 1)$ generates training dynamics that are superior to the baseline $\tvo$ and to $\ovismc$ given a comparable particle budget (appendix \ref{apdx:budget}). 
We interpret this result as a consequence of the $\ess$-specific design, which also appeared to be robust to the choice of $\alpha$ in the IWR objective. This also corroborates the results of~\cite{Roeder2017-qc}, that suppressing the term $- \sum_k \vk \hk$ from the gradient estimate improves learning. We therefore recommend the practitioner to first experiment with $\ovisinf(\gamma=1)$ since it delivers competitive results at a reasonable computational cost.

\section{Conclusion}

We proposed $\ovis$, a gradient estimator that is generally applicable to deep models with stochastic variables, and is empirically shown to have optimal variance control. This property is achieved by identifying and canceling terms in the estimator that solely contribute to the variance. We expect that in practice it will often be a good trade-off to use a looser bound with a higher effective sample size, e.g. by utilizing the $\ovis$ estimator with the importance weighted Rényi bound, allowing control of this trade-off via an additional scalar smoothing parameter. This sentiment is supported by our method demonstrating better performance than the current state-of-the-art.


\section{Financial Disclosure}

The PhD program supporting Valentin Liévin is partially funded by Google. This research was supported by the NVIDIA Corporation with the donation of GPUs.

\section{Broader Impact}
This work proposes OVIS, an improvement to the score function gradient estimator in the form of optimal control variates for variance reduction. As briefly touched upon in the introduction, OVIS has potential practical use cases across several branches of machine learning. As such, the potential impact of this research is broad, and we will therefore limit the scope of this section to a few clear applications.

Improved inference over discrete spaces such as action spaces encountered within e.g. model-based reinforcement learning has the potential of reducing training time and result in more optimal behavior of the learning agent. This advancement has the capability to increase efficiency of e.g. autonomous robots used within manufacturing. Such progress is often coveted due to cost optimization, increased safety, and reduced manual labor for humans. However, as argued in \citep{groover_2019}, this development can also lead to immediate disadvantages such as worker displacement, potentially in terms of both tasks and geographic location. 

Another probable avenue of impact of this research is within machine comprehension. A topic within this field is reading, with practical applications such as chatbots. This use of machine learning has seen rapid growth and commercial interest over recent years \citep{nguyen_2020}. Apart from the clear consumer benefits of these bots, focus has also broadened to other cases of use for social benefits \citep{folstad_sig}. However, as with most other machine learning inventions, chatbots can be exploited for malicious purposes such as automated spread of misinformation, e.g. during elections \citep{matthews_2019}. 


As with other theoretical advances such as those presented in this paper, consequences are not immediate and depend on the applications in which the research is utilized. It is our hope that this research will ultimately be of practical use with a tangible positive impact.  

\bibliographystyle{unsrtnat}
\bibliography{main}

\newpage
\appendix 

\section{Derivation of the Score Function Estimator}\label{apdx:grads}

Given $K$ samples, the objective being maximized is
%
\begin{align}
    \LK(\xb) \defeq \Ex
    \left[ \log \ZK \right] \qquad \ZK \defeq \iwf \qquad \wk \defeq \frac{\pth(\xb, \zb_k)}{\qphi(\zb_k | \xb)} \ .
\end{align}
The gradients of the multi-sample objective $\LK$ with respect to the parameter $\phi$ can be expressed as a sum of two terms, one arising from the expectation over the variational posterior $\qphizs \defeq \prod_{k=1}^{K} \qphizk$ and one from $\log \ZK$: 
\begin{align}
    \gphi \LK 
    & = \underbrace{ 
        \Ex \left[ \log \ZK  \frac{\gphi \qphizs}{\qphizs} \right]
    }_\text{\textbf{(a)}} + 
    \underbrace{ 
        \Ex \left[ \gphi \log \ZK \right]
    }_\text{\textbf{(b)}} \nonumber \ .
\end{align}
The term \textbf{(a)} yields the traditional score function estimator 
\begin{align}
    \text{\textbf{(a)}} 
    & = \Ex
    \left[ \log \ZK \, \gphi \log \qphizs \right] \nonumber\\
    & = \Ex
    \left[\log \ZK \, \sum_{k=1}^K    \gphi \log \qphizk \right] \label{eq:reinforce} \ .
\end{align}
The term \textbf{(b)} is 
\begin{align}
    \text{\textbf{(b)}} 
    & = \Ex
    \left[  \gphi \log \iwf \right] \nonumber\\
    & = \Ex
    \left[  \frac{1}{\expiwf} \gphi \expiwf \right] \nonumber\\
    & = \Ex
    \left[ \frac{1}{\sum_{l=1}^K \wl} \sum_{k=1}^K \gphi \wk \right] \nonumber\\
     & = \Ex
    \left[ \frac{1}{\sum_{l=1}^K \wl} \sum_{k=1}^K \wk \gphi \log \wk \right] \nonumber\\
    & =  \Ex
    \left[ \sum_{k=1}^K \vk \gphi \log \wk \right], \quad \vk = \frac{\wk}{\sum_{l=1}^K \wl} \nonumber
    \\
     & = -\Ex
     \left[ \sum_{k=1}^K \vk   \gphi \log \qphizk \right]  \ .
\end{align}
The derivation yields a factorized expression of the gradients
\begin{align}
    \gphi \LK & = \Ex_{\qphizs} \left[\sum_{k=1}^K  \left(\log \ZK - \vk \right) \hk \right] \quad \text{ with }  \quad \hk   \defeq \gphi \log \qphizk \ .
\end{align}

\section{Asymptotic Analysis}\label{app:asymptotic_analysis}

We present here a short derivation and direct the reader to \cite{Rainforth2018-pw} for the fine prints of the proof. The main requirement is that $w_k$ is bounded, so that $\ZK - Z$ (with $Z=p(\xb)$) will converge to 0 almost surely as $K\to\infty$. We can also state this through the central limit theorem by noting that $\ZK - Z=\frac{1}{K} \sum_k (w_k - Z)$ is the sum of $K$ independent terms so if $\Var_1[w_1]$ is finite then $\ZK - Z$ will converge to a Gaussian distribution with mean $\Ex[\ZK - Z]=0$ and variance $\Var[\ZK - Z]=\frac{1}{K}\Var_1[w_1]$. The $K^{-1}$ factor on the variance follows from independence. This means that in a Taylor expansion in $\ZK - Z$ higher order terms will be suppressed.

Rewriting $\grad$ in terms of $\log \ZK$:
\begin{align}
    \grad 
    = \sum_k \left(\gk - \ck \right) \hk 
    = \sum_k \left( \log \ZK -  w_k \frac{\partial }{\partial w_k} \log \ZK - \ck \right) \hk
\end{align}
and using the second-order Taylor expansion of $\log \ZK$ about $Z$:
\begin{equation}
    \log \ZK \approx \log Z + \frac{\ZK - Z }{Z} - \frac{(\ZK - Z)^2}{2Z^2}    
\end{equation}
we have
\begin{gather}
    \log \ZK \approx \log Z - \frac{3}{2} + \frac{2}{KZ} \sum_l \wl - \frac{1}{2 K^2 Z^2} \left( \sum_l \wl \right)^2 \\
    \frac{\partial }{\partial w_k} \log \ZK \approx \frac{2}{KZ} - \frac{1}{K^2 Z^2} \sum_l \wl \  .
\end{gather}
The term $\gk$ can thus be approximated as follows:
\begin{align}
    \gk 
    &=  \log \ZK -  w_k \frac{\partial }{\partial w_k} \log \ZK \nonumber  \\
    & \approx \log Z - \frac{3}{2} + \frac{2}{KZ} \sum_{l \neq k} \wl - \frac{1}{2 K^2 Z^2} \left( \sum_l \wl \right)^2 + \frac{1}{K^2 Z^2} w_k \sum_l \wl \nonumber \\
    &= \log Z - \frac{3}{2} + \frac{2}{KZ} \sum_{l \neq k} \wl - \frac{1}{2 K^2 Z^2} \Bigg( \sum_{l \neq k} \wl \Bigg)^2 + \frac{1}{2 K^2 Z^2} \wk^2 \label{eq:app:approx_dk}
\end{align}
where we used
\begin{align*}
    \Bigg( \sum_l \wl \Bigg)^2 = \Bigg( \sum_{l \neq k} \wl \Bigg)^2 + \wk^2 + 2 \wk \sum_{l \neq k} \wl \ .
\end{align*}
By separately collecting the terms that depend and do not depend on $\zb_k$ into $\fk=\fk(\zb_k,\zb_{-k})$ and $\fnok=\fnok(\zb_{-k})$, respectively, we can rewrite the estimator $\grad$ as:
\begin{equation}
    \grad = \sum_k (f_k + f_{-k} - c_k)\hk
\end{equation}
and from~\eqref{eq:app:approx_dk} we have
\begin{gather}
    f_k \approx \frac{\wk^2}{2 K^2 Z^2} \label{eq:appendix:fk} \\
    f_{-k} \approx \log Z - \frac{3}{2} + \frac{2}{KZ} \sum_{l \neq k} \wl - \frac{1}{2 K^2 Z^2} \Bigg( \sum_{l \neq k} \wl \Bigg)^2 \ .
\end{gather}

\section{Asymptotic Expectation and Variance}

We derive here the asymptotic expectation and variance of the gradient estimator $\grad$ in the limit $K \to \infty$.

\subsection{Expectation}\label{apds:asymptoticgrad:expectation}
If both $\fnok$ and $\ck$ are independent of $\zb_k$, we can write:
\begin{align}
    \Ex [\grad] = \Ex \left[ \sum_k (f_k + f_{-k} - c_k)\hk \right] = \sum_k \Ex \left[\fk \hk \right]
\end{align}
where we used that $\Ex \left[\fnok \hk \right]$ and $\Ex \left[\ck \hk \right]$ are zero. 
In the limit $K \to \infty$, each term of the sum can be expanded with the approximation~\eqref{eq:appendix:fk} and simplified:
\begin{align}
    \Ex \left[\fk \hk \right] 
    & \approx  \Ex \left[ \frac{\wk^2}{2 K^2 Z^2} \hk \right] 
    =  \frac{1}{2 K^2 Z^2} \Ex_1 \left[ \wone^2 \hone \right]
\end{align}
where $\Ex_1$ denotes an expectation over the posterior $\qphizX{1}$. The last step follows from the fact that the latent variables $\{\zb_k\}_{k=1}^K$ are i.i.d. and the argument of the expectation only depends on one of them. In conclusion, the expectation is:
\begin{align}
    \Ex [\grad] = \sum_k \Ex \left[\fk \hk \right] \approx \frac{1}{2 K Z^2} \Ex_1 \left[ \wone^2 \hone \right] = \order(K^{-1})
\end{align}
irrespective of $\fnok$ and $\ck$.

\subsection{Variance}\label{apds:asymptoticgrad:variance}
If $\ck$ is chosen to be $\ck(\znok) = \fnok(\znok)$ then we can again use the approximation~\eqref{eq:appendix:fk} for $K \to \infty$ and get the asymptotic variance:
\begin{align}
    \Var [\grad] & = \Var \left[ \sum_k \fk \hk \right] \\
    &\approx \Var \left[ \sum_k \frac{\wk^2}{2 K^2 Z^2} \hk \right] \\
    & = \frac{1}{4 K^4 Z^4} \sum_k \Var_k \left[ \wk^2 \hk \right] \\
    & = \frac{1}{4 K^3 Z^4} \Var_1 \left[ \wone^2 \hone \right] \\
    & = \order(K^{-3})
\end{align}
where $\Var_k$ denotes the variance over the $k$th approximate posterior $\qphizX{k}$, and we used the fact that the latent variables $\{\zb_k\}_{k=1}^K$ are i.i.d. and therefore there are no covariance terms.

\section{Optimal Control for the ESS Limits and Unified Interpolation}\label{apds:ESS}

\subsection{Control Variate for Large ESS}\label{apds:ESSlarge}

In the gradient estimator $\grad = \sum_k \left( \log \ZK  -  \frac{\partial \log \ZK}{\partial w_k} w_k - \ck \right) \hk$, we consider the $k$th term in the sum, where we have that $\ZK - \ZKnok = \frac{\wk}{K} \to 0$ as $K \to \infty$. We can therefore expand $\log \ZK$ as a Taylor series around $\ZK=\ZKnok$, obtaining:
\begin{align}
\log \ZK & = \log \ZKnok + \sum_{p=1}^\infty \frac{(-1)^{p+1}}{p} \left( \frac{\wk}{K \ZKnok} \right)^p \\
\frac{\partial \log \ZK}{\partial w_k} & = \frac{1}{\wk} \sum_{p=1}^\infty (-1)^{p+1} \left( \frac{\wk}{K \ZKnok} \right)^p
\ .  
\end{align}
Inserting these results into the gradient estimator and using the expression $\grad = \sum_k (f_k + f_{-k} - c_k)\hk$ we see that
\begin{align}
\fnok &= \log \ZKnok \\
\fk &= \sum_{p=1}^\infty (-1)^{p+1} \left(\frac{1}{p} - 1 \right) \left( \frac{\wk}{K \ZKnok} \right)^p \\
    &= \sum_{p=2}^\infty (-1)^{p} \left( 1 - \frac{1}{p} \right) \left( \frac{\wk}{K \ZKnok} \right)^p 
    \ .  
\end{align}
We now use this to simplify the optimal control variate~\eqref{eq:copt} to leading order. Since $\fk$ is order $K^{-2}$, the term $\Ex_k\left[ \fk \normhk^2 \right]$ will be of order $K^{-2}$ as well. The $l\neq k$ terms $\Ex_k\left[ \fl \hk^T \hl \right]$ get non-zero contributions only through the $\wk$ term in $\fl$. As $\wk$ appears in $\ZKnol$ with a prefactor $K^{-1}$, we have $\Ex_k\left[ \fl \hk^T \hl \right] = \order(K^{-3})$ for $l\neq k$, and the sum of these terms is $\order(K^{-2})$. Overall, this means that the second term in the control variate only gives a contribution of $\order(K^{-2})$ and thus can be ignored:
\begin{equation}
\ck \approx \log \ZKnok = \log \frac{1}{K} \sum_{l\neq k} \wl = \log \frac{1}{K-1} \sum_{l\neq k} \wl + \log(1 - \frac{1}{K}) \ .
\end{equation}
Note that in the simplifying approximation in Section~\ref{sec:optimal_control} we argue that the $l\neq k$ terms $\Ex_k\left[ \fl \hk^T \hl \right]$ can be omitted and only the $l=k$ term retained. Here we show that their overall contribution is the same order as the $l=k$ term. These results are not in contradiction because here we are only discussing orders and not the size of terms. 

\subsection{Control Variate for Small ESS}\label{apds:ESS1}

In the case $\ess \approx 1$ we can write $\log \ZK$ as a sum of two terms:
\begin{equation}
    \log \ZK = \log \frac{w_{k'}}{K} + \log \left( 1 + \frac{K \tilde{Z}_{[-k']}}{w_{k'}} \right) \ ,
\end{equation}
where $w_{k'}$ is the dominating weight.
%
The first term dominates and the second can be ignored to leading order. We will leave out a derivation for non-leading terms for brevity. So the gradient estimator $\grad = \sum_k \left(\log \ZK  -  \frac{\partial \log \ZK}{\partial w_k} w_k - \ck \right) \hk$ simply becomes $\grad \approx \sum_k \left(  \log \frac{w_{k'}}{K} - \delta_{k,k'} - \ck \right) \hk$. This corresponds to $f_k = \delta_{k,k'} \log w_{k'}$ and $f_{-k}=(1-\delta_{k,k'} )\log w_{k'} - \delta_{k,k'} - \log K$. Inserting this into Equation~\eqref{eq:gradopt} we get:
\begin{equation}
    \grad = \sum_k \left(\fk - \sum_l \frac{\Ex_k \left[ \fl \hk^T \hl \right]}{\Ex_k \left[ \normhk^2   \right]} \right) \hk 
    = \left( \log w_{k'}  - \frac{\Ex_{k'} \left[ \log w_{k'} ||\mathbf{h}_{k'}||^2 \right]}{\Ex_{k'} \left[ ||\mathbf{h}_{k'}||^2   \right]} \right) \mathbf{h}_{k'} \ . \label{eq:grad-ess-small}
\end{equation}

Estimating the expectation $\Ex_{k'} [ \dots ]$ in Equation~\eqref{eq:grad-ess-small} using i.i.d. samples from $\qphi(\zb | \xb)$ is computationally involved. Therefore we resort to the approximation  $\grad \approx \sum_k \left(  \log \frac{w_{k'}}{K} - \delta_{k,k'} - \ck \right) \hk$  and $\delta_{k,k'} \approx \vk$, which holds in the limit $\ess \rightarrow 1$. We get:
\begin{equation}
    \ck \approx \log \ZKnokunb - \vk = \log \frac{1}{K-1} \sum_{l \neq k} \wl - \vk \ .
\end{equation}

Relying on the approximation $\delta_{k,k'} \approx \vk$ corresponds to suppressing the term $-\vk$ of the prefactors $\gk = \log \ZK - \vk$ and does not guarantee the resulting objective to be unbiased for $\ess > 1$. Suppressing this term has been explored in depth for the pathwise gradient estimator~\cite{Roeder2017-qc}. The gradient estimator $\sum_k \vk \hk$ corresponds to \textit{wake-phase} update in $\rws$.

\subsection{Unified Interpolation}\label{apds:unified-control}

We unify the two $\ess$ limits under a unifying expression $\ovisinf$ defined for a scalar $\gamma \in [0, 1]$:
\begin{equation}
    \ck^{\gamma} \defeq \log \ZKnokunb - \gamma  \vk + (1 - \gamma)  \log(1 - 1/K) 
\end{equation}
where
\begin{align}
    \ck^0 & = \log \frac{1}{K-1} \sum_{l \neq k} \wl + \log (1-1/K) \\
    \ck^1 & = \log \frac{1}{K-1} \sum_{l \neq k} \wl - \vk \ .
\end{align}

\section{Rényi Importance Weighted Bound}\label{apdx:beta-bound}

All the analysis applied to the score function estimator for the importance weighted bound including asymptotic $\snr$ can directly be carried over to the Rényi importance weighted bound $\LK^{\alpha}(\xb)$ because all the independence properties are unchanged. The score function estimator of the gradient of $\phi$ is given by 
\begin{align}
    \gphi \LK^{\alpha}(\xb) & = \sum_k \left( \frac{1}{1-\alpha} \log \ZK(\alpha) - \vk(\alpha) \right) \hk, \qquad \vk(\alpha) = \frac{\wk^{1 - \alpha}}{\sum_l \wl^{1 - \alpha}} \ .
\end{align}
The $\ovismc$ formulation holds using $d_k =  \frac{1}{1-\alpha} \log \ZK(\alpha) - \vk(\alpha)$ within the equation~\ref{eq:sample-based-control}. Similarly for the asymptotic expression $\ovisinf$, the unified control variate~\ref{eq:alpha-control} becomes:
\begin{gather}
\quad \ck^{\gamma} \defeq \log \frac{1}{1-\alpha} \log \ZKnokunb(\alpha) - \gamma  \vk + (1 - \gamma) \log(1 - 1/K)
\end{gather}

\section{Gradient Estimators Review}\label{apdx:estimators}
In this paper, gradient \emph{ascent} is considered (i.e. maximizing the objective function). The expression of the gradient estimators presented below are therefore adapted for this setting. 

\paragraph{VIMCO} The formulation of the $\vimco$~\citep{Mnih2016-vr} control variate exploits the structure of $\ZK \defeq \frac{1}{K} \sum_l \wl$ using $\ck \defeq \ck(\znok) = \log \frac{1}{K} \sum_{l \neq k} \wl + \wnok$ where $\wnok$ stands for the arithmetic or geometric average of the weights $\wl$ given the set of outer samples $\znok$. Defining $\log \ZKnokunb \defeq \ck$, the $\vimco$ estimator of the gradients is
\begin{align}
    \gphi \LK  = \Ex_{\qphizs} \Bigg[ 
        \underbrace{ \sum_{k=1}^K \left(\log \ZK - \log \ZKnokunb \right) \hk }_\text{\textbf{(a)}}
        + 
        \underbrace{ \sum_{k=1}^K \vk \gphi \log \wk }_\text{\textbf{(b)}}
    \Bigg] \ .
\end{align}

We refer to \citep{Mnih2016-vr} for the derivation. Here, the term $\ZKnokunb$ can be expressed using the arithmetic and the geometric averaging~\citep{Mnih2016-vr}. The leave-one-sample estimate can be expressed as
\begin{align}
    \ZKnokunb = \frac{1}{K} \sum_{l \neq k} \wl + \wnok \text{ with } 
    \begin{cases} 
    \wnok = \frac{1}{K-1} \sum_{l \neq k} \wl & \text{(arithmetic)} \\
    \wnok = \exp{ \frac{1}{K-1} \sum_{l \neq k} \log \wl} & \text{(geometric)}
    \end{cases}
\end{align}
The term \textbf{(b)} is well-behaved because it is a convex combination of the K gradients $\gphi \log \wk $. However, the term \textbf{(a)} may dominate the term \textbf{(b)}. In contrast to $\vimco$, $\ovis$ allows controlling the variance of both terms \textbf{(a)} and \textbf{(b)}, resulting in a more optimal variance reduction. In the Reweighted Wake Sleep ($\rws$) with wake-wake-$\phi$ update, the gradient of the parameters $\phi$ of the inference network corresponds to the negative of the term \textbf{(b)}.

\paragraph{Wake-sleep} The algorithm~\citep{Hinton1995-qk} relies on two separate learning steps that are alternated during training: the \textit{wake-phase} that updates the parameters of the generative model $\theta$ and the \textit{sleep-phase} used to update the parameters of the inference network with parameters $\phi$. During the \textit{wake-phase}, the generative model is optimized to maximize the evidence lower bound $\mathcal{L}_1$ given a set of observation $\xb \sim p(\xb)$. During the \textit{sleep-phase}, a set of observations and latent samples are \textit{dreamed} from the model: $\xb,\zb \sim \pth(\xb, \zb)$ and the parameters $\phi$ of the inference network are optimized to minimize the $\operatorname{KL}$ divergence between the true posterior of the generative model and the approximate posterior: $\kl \left( \pth (\zb | \xb) || \qphi (\zb | \xb)  \right)$.

\paragraph{Reweighted Wake-Sleep (RWS)} extends the original Wake-Sleep algorithm for importance weighted objectives~\citep{Bornschein2014-my}. The generative model is now optimized for the importance weighted bound $\LK$, which gives the following gradients
\begin{equation}
    \nabla_{\theta} \LK = \Ex_{\qphizs} \left[  \sum_k \vk \nabla_{\theta} \log \wk \right] \quad \text{(wake-phase $\theta$)} \ .
\end{equation}
The parameters $\phi$ of the inference network are optimized given two updates: the \textit{sleep-phase $\phi$} an the \textit{wake-phase $\phi$}. The \textit{sleep-phase} $\phi$ is identical to the original Wake-Sleep algorithm, the gradients of the parameters $\phi$ of the inference model are given by
\begin{equation}
    - \gphi \Ex_{\pth(\xb)} \left[ \kl \left( \pth (\zs | \xb) || \qphi (\zs | \xb)  \right) \right]
    = \Ex_{\pth(\zs, \xb)} \left[ \sum_k \hk \right] \quad \text{(sleep-phase $\phi$)} \ .
\end{equation}
The \textit{wake-phase} $\phi$ differs from the original Wake-Sleep algorithm that samples $\xb, \zb$ are sampled respectively from the dataset and from the inference model $\qphi(\zb|\xb)$. In this cases the gradients are given by:

\begin{equation}
    - \gphi \Ex_{p(\xb)} \left[ \kl \left( \pth (\zs | \xb) || \qphi (\zs | \xb)  \right) \right]  
    = \Ex_{p(\xb)} \left[ \Ex_{\qphizs} \left[  \sum_k \vk \hk \right] \right] \quad \text{(wake-phase $\phi$)} \ .
\end{equation}
Critically, in Variational Autoencoders one optimizes a lower bound of the marginal log-likelihood ($\LK$), while RWS instead optimizes a biased estimate of the marginal log-likelihood $\log p(\xb)$. However, the bias decreases with $K$~\citep{Bornschein2014-my}. ~\cite{Le2018-kj} shows that RWS is a method of choice for training deep generative models and stochastic control flows. In particular, ~\cite{Le2018-kj} shows that increasing the budget of particles $K$ benefits the learning of the inference network when using the wake-phase update (Wake-Wake algorithm).

We refer the reader to~\cite{Bornschein2014-my} for the derivations of the gradients and~\cite{Le2018-kj} for an extended review of the RWS algorithms for the training of deep generative models.

\paragraph{The Thermodynamic Variational Objective (TVO)} The gradient estimator consists of expressing the marginal log-likelihood $\log \pth(\xb)$ using Thermodynamic Integration (TI). 
Given two unnormalized densities $\tilde{\pi}_0(\zb)$ and $\tilde{\pi}_1(\zb)$ and their respective normalizing constants $Z_0, Z_1$ with $Z_i = \int \tilde{\pi}_i(\zb) d \zb$ given the unnormalized density 
$\tilde{\pi}_{\beta}(\zb) \defeq \pi_1(\zb)^{\beta} \pi_{0}^{1 - \beta}(\zb)$ parameterized by $\beta \in [0,1]$, and the corresponding normalized density $\pi_{\beta}(\zb) = \tilde{\pi}_{\beta}(\zb) / \int \tilde{\pi}_{\beta}(\zb) d\zb$, TI seeks to evaluate the ratio of the normalizing constants using the identity 
\begin{equation}
\log Z_1 - \log Z_0 = \int_0^1 \Ex_{\pib} \left[ \frac{d \log \upib (\zb)}{d \beta} \right] d\beta \ .
\end{equation}

~\cite{Masrani2019-nq} connects TI to Variational Inference by setting the base densities as $\upi_0(\zb) = \qphi(\zb | \xb)$ and $\upi_1(\zb) = \pth(\xb, \zb)$, which gives the Thermodynamic Variational Identity (TVI):

\begin{equation}
\log p_{\theta}(\mathbf{x}) = \int_{0}^{1} \mathbb{E}_{\pi_{\beta}}\left[\log \frac{p_{\theta}(\mathbf{x}, \mathbf{z})}{q_{\phi}(\mathbf{z} | \mathbf{x})}\right] d \beta.
\end{equation}

Applying left Riemannian approximation yields the Thermodynamic Variational Objective ($\tvo$):

\begin{equation}
\tvo(\theta, \phi, \xb) = \frac{1}{P} \left[\elbo(\theta, \phi, \mathbf{x})+\sum_{p=1}^{P-1} \mathbb{E}_{\pi_{\beta_{P}}}\left[\log \frac{p_{\theta}(\mathbf{x}, \mathbf{z})}{q_{\phi}(\mathbf{z} | \mathbf{x})}\right]\right] \leq \log p_{\theta}(\mathbf{x}) \ .
\end{equation}

Notably, the integrand $\Ex_{\pib} \left[ \log \frac{\pth(\xb, \zb)}{\qphi(\zb | \xb)} \right]$ is monotically increasing, which implies that the $\tvo$ is a lower-bound of the marginal log-likelihood. 

The $\tvo$ allows connecting both Variational Inference and the Wake-Sleep objectives by observing that when using a partition of size $P=1$, the left Riemannian approximation of the TVI, $\tvo_1^{L} (\theta, \phi, \xb) = \elbo (\theta, \phi, \xb)$ and the right Riemannian approximation of the TVI, $\tvo_1^U (\theta, \phi, \xb) $ is an upper bound of the marginal log-likelihood and equals the objective being maximized in the \textit{wake-phase} for the parameters $\phi$ of the inference network.

Estimating the gradients of the $\tvo$ requires computing the gradient for each of the $P$ expectations $\Ex_{\pi_{\lambda, \beta}} \left[ f_{\lambda} (\zb) \right]$ with respect to a parameter $\lambda \defeq \{ \theta, \phi \}$ where $f_{\lambda} (\zb) = \log \frac{\pth(\xb, \zb)}{\qphi(\zb | \xb)} $ and $\xb$ is fixed. In the general case, differentiation through the expectation is not trivial. Therefore the authors propose a score function estimator
\begin{equation}
    \nabla_{\lambda} \mathbb{E}_{\pi_{\lambda, \beta}}\left[f_{\lambda}(\mathbf{z})\right] = \Ex_{\pi_{\lambda, \beta}}\left[\nabla_{\lambda} f_{\lambda}(\mathbf{z})\right]+\Cov_{\pi_{\lambda, \beta}}\left[\nabla_{\lambda} \log \tilde{\pi}_{\lambda, \beta}(\mathbf{z}), f_{\lambda}(\mathbf{z})\right] \ ,
\end{equation}
where the covariance term can be expressed as
\begin{equation}
    \Ex_{\pi_{\lambda, \beta}} \left[
    \left( f_{\lambda}(\zb) -  \Ex_{\pi_{\lambda, \beta}}\left[ f_{\lambda}(\mathbf{z}) \right] \right)
    \left( \nabla_{\lambda} \log \tilde{\pi}_{\lambda, \beta}(\mathbf{z}) -  \Ex_{\pi_{\lambda, \beta}}\left[ \nabla_{\lambda} \log \tilde{\pi}_{\lambda, \beta}(\mathbf{z}) \right] \right)
    \right] \ .
\end{equation}

The covariance term arises when differentiating an expectation taken over a distribution with an intractable normalizing constant, such as  $\pib(\zb)$ in the TVO. The normalizing constant can be substituted out, resulting in a covariance term involving the tractable un-normalized density $\upib(\zb)$. Hence, such a covariance term does not usually arise in IWAE due to the derivative of $\qphi(\zb | \xb)$ being available in closed form.

\section{Gaussian Model}\label{apdx:fit-gaussian}

\paragraph{Distribution of gradients}

\begin{figure}[h]
  \centering
  \includegraphics[width=1.\textwidth]{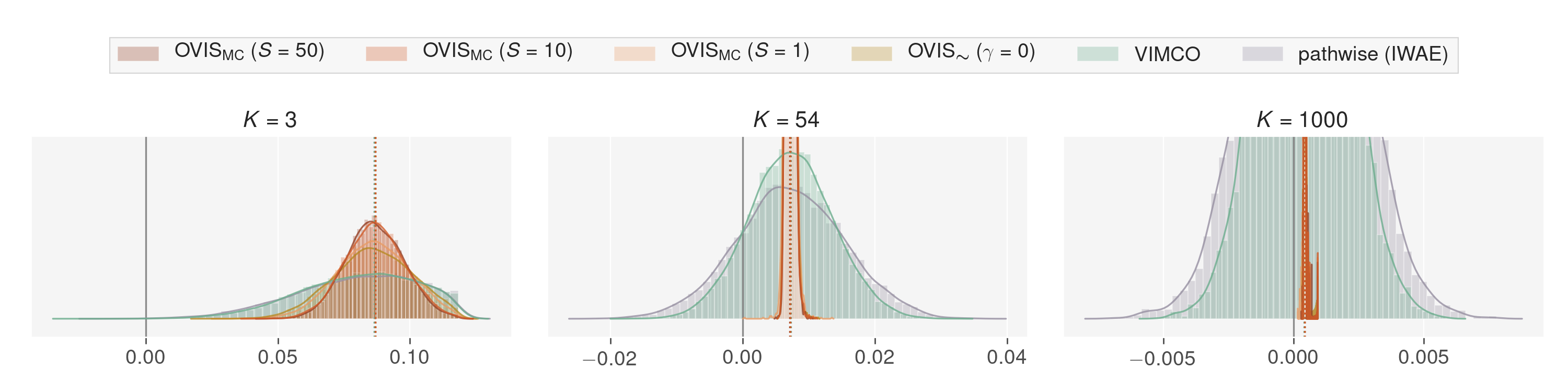}
  \caption{Distribution of the gradients for an arbitrarily chosen component of the parameter $\mathbf{b}$. The tight control of the variance provided by $\ovis$ allows keeping the distribution of gradients off-center.}\label{fig:asymptotic-variance-dist}
\end{figure}

We report the distributions of the $10^{4}$ MC estimates of the gradient of the first component $b_0$ of the parameter $\mathbf{b}$. Figure~\ref{fig:asymptotic-variance-dist}. The pathwise estimator and $\vimco$ yield estimates which distributions are progressively centered around zero as $K \to \infty$. The faster decrease of the variance of the gradient estimate for $\ovis$ results in a distribution of gradients that remains off-centered.

\paragraph{Analysis for advanced pathwise IWAE estimators}

\begin{figure}[h]
  \centering
  \includegraphics[width=1.\textwidth]{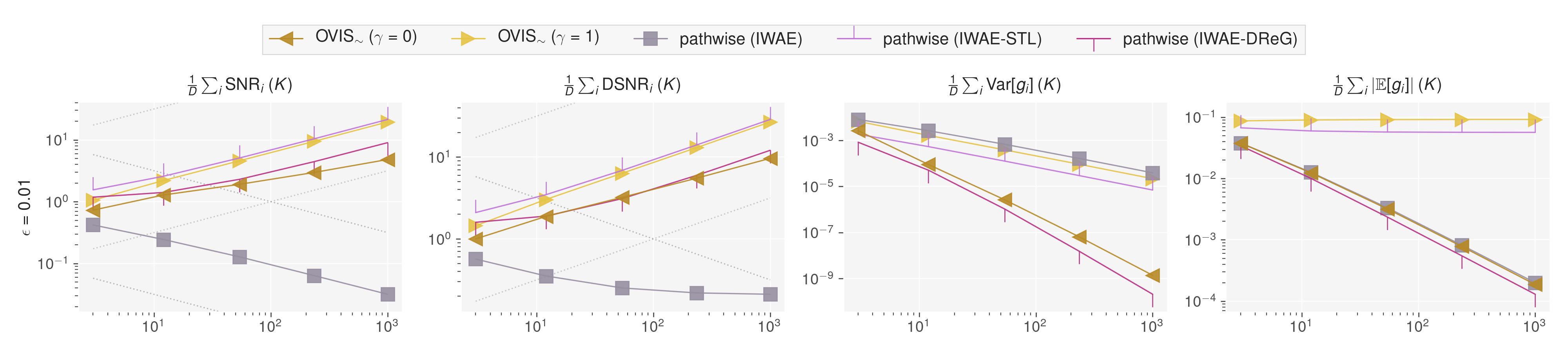}
  \caption{Asymptotic analysis of the gradients for $\ovisinf$ and the STL and DReG IWAE estimators.}\label{fig:asymptotic-variance-advanced}
\end{figure}

We perform the experiment \ref{sec:asymptotic} using additional pathwise estimators: STL \citep{Roeder2017-qc} and DReG-IWAE \citep{Tucker2018-nr}. Both the STL and $\ovisinf (\gamma = 1)$ rely on the suppression of the term $\sum_k \vk \hk$ from the gradient estimate and adopt the same behaviour: the variance decreases at a slower rate than $\ovisinf (\gamma=0)$ and DReG, however, its bias remains constant as K is increased.

\newpage
\paragraph{Fitting the Gaussian Model}

\begin{center}
  \captionsetup{type=figure}
   \includegraphics[width=1\textwidth]{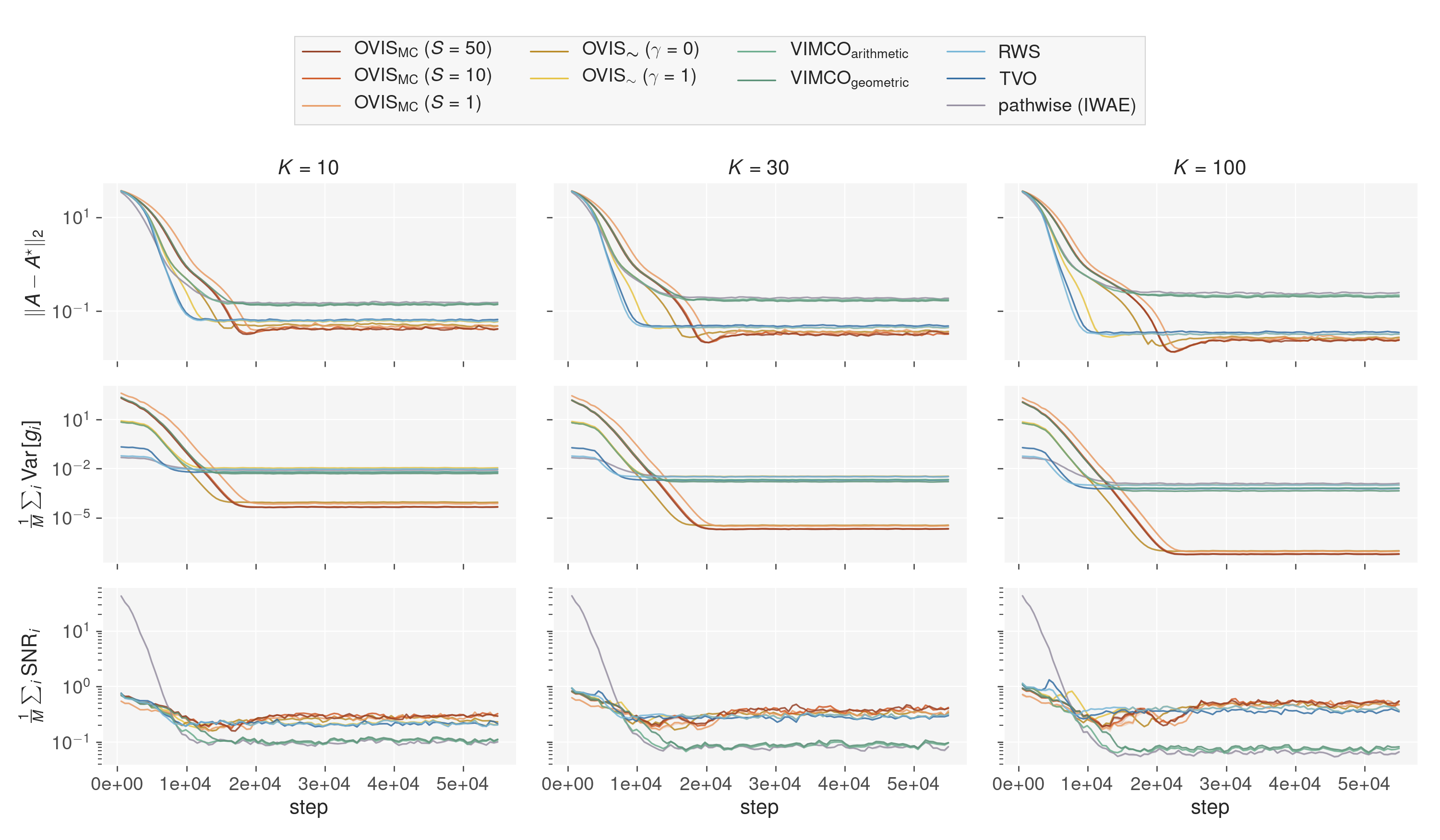}
   \captionof{figure}{ Fitting the Gaussian toy model from section~\ref{exp:asymptotic-variance} and measuring the $\mathcal{L}_2$ distance with the optimal parameters as well as the variance and the $\snr$ of the gradient estimates. $\ovis$ methods target the optimal parameters $A^{\star}$ of the inference network more accurately than the baseline methods.}\label{fig:fitting-gaussian-toy}
 \end{center}

We study the relative effect of the different estimators when training the Gaussian toy model from section ~\ref{exp:asymptotic-variance}. The model is trained for 5.000 epochs using the Adam optimizer with a base learning rate of $10^{-3}$ and with a batch-size of 100. In Figure~\ref{fig:fitting-gaussian-toy}, we report the $L_2$ distance from the model parameters $A$ to the optimal parameters $A^\star$, the parameters-average $\snr$ and parameters-average variance of the inference network $(\phi = \{A, \mathbf{b}\}, M = \operatorname{card}(\phi))$. We compare $\ovis$ methods with $\vimco$, the pathwise $\iwae$, $\rws$ and the $\tvo$ for which we picked a partition size $P=5$ and $\beta_1 = 10^{-3}$, although no extensive grid search has been implemented to identify the optimal choice for this parameters.

$\ovis$ yields gradient estimates of lower variance than the other methods. The inference network solutions given by $\ovis$ are slightly more accurate than the baseline methods $\rws$ and the $\tvo$, despite being slower to converge. $\ovis$, $\rws$ and the $\tvo$ exhibit gradients with comparable $\snr$ values, which indicate $\ovis$ yield estimate of lower expected value, thus leading to a smaller maximum optimization step-size. Setting $\gamma = 0$ for $\ovisinf$ results in more accurate solutions than using $\gamma = 1$, this coincides with the measured $\ess \approx K$.


\newpage
\section{Gaussian Mixture Model}\label{apdx:gmm}


\begin{center}
  \captionsetup{type=figure}
  \includegraphics[width=\textwidth]{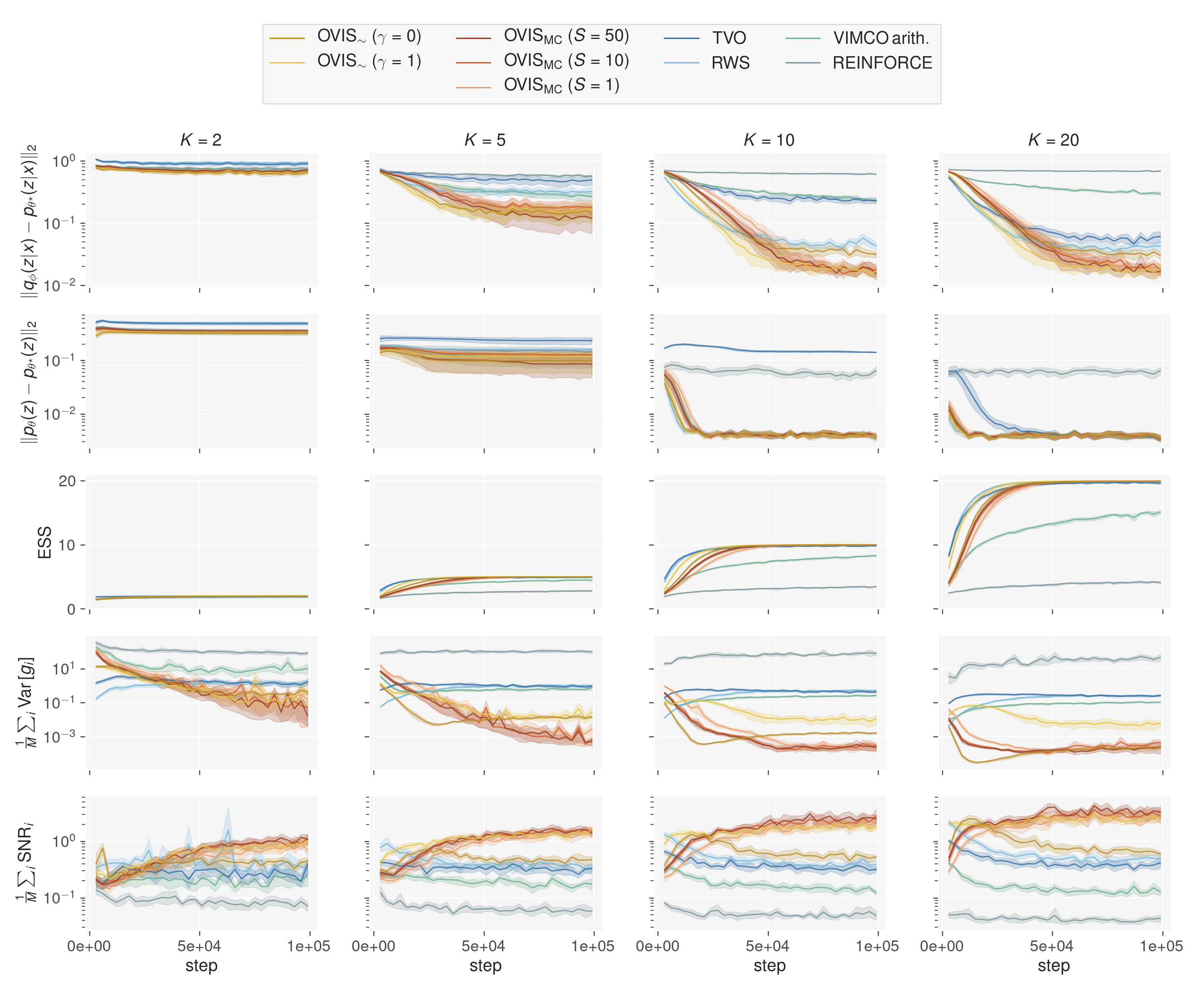}
  \captionof{figure}{Training curves for the Gaussian Mixture Model for different numbers of particles $K = [2, 5, 10, 20]$ samples averaged over 5 random seeds. The $\snr$ is measured on one mini-batch and averaged over the $M$ parameters of the inference network. In contrast to $\vimco$, $\ovis$ estimators all generate gradients with a higher $\snr$. This results in a more accurate estimate of the true posterior, when compared to $\vimco$ and the baselines $\rws$ and the $\tvo$.}\label{fig:gmm-detailed}
\end{center}

\section{Comparison of $\ovisinf$ and $\ovismc$ with under a fixed Particle Budget}\label{apdx:budget}

$\ovismc$ has complexity requires $K+S$ importance weights whereas $\ovisinf$ requires only $K$. Estimating $\phi$ using $\ovismc$ requires a budget of $K' = K + S$ particles. The ratio $S / K$ is a trade-off between the tightness of the bound $\LK$ and the variance of the control variate estimate. In the main text, we focus on studying the sole effect of the control variate given the bound $\LK$. This corresponds to a sub-optimal use of the budget $K'$ because $\mathcal{L}_{K'}$ is tighter than $\mathcal{L}_{K}$. By contrast with the previous experiments, we trained the Gaussian VAE using the budget $K'$ optimally (i.e. relying on $\mathcal{L}_{K'}$ whenever no auxiliary samples are used). We observed that $\ovisinf (\gamma = 1)$ outperforms $\ovismc$ despite the generative model is evaluated using $\mathcal{L}_{K'}$ in all cases (figure \ref{fig:budget}). This experiment will be detailed in the Appendix.

\begin{center}
  \captionsetup{type=figure}
        \includegraphics[width=1.\linewidth]{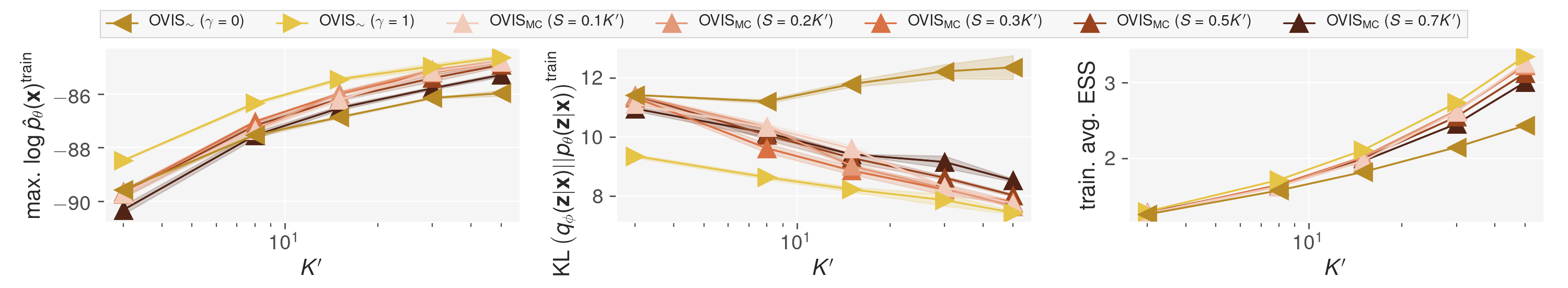}
  \captionof{figure}{Training the Gaussian VAE model with a fixed and optimally used particle budget $K' = K+S$ and $\alpha=0.7$.}\label{fig:budget}
\end{center}


\section{Training Curves for the Deep Generative Models}\label{apdx:dgm}

\subsection{Sigmoid Belief Network}\label{apdx:sbm}

\begin{center}
  \captionsetup{type=figure}
  \includegraphics[width=\textwidth]{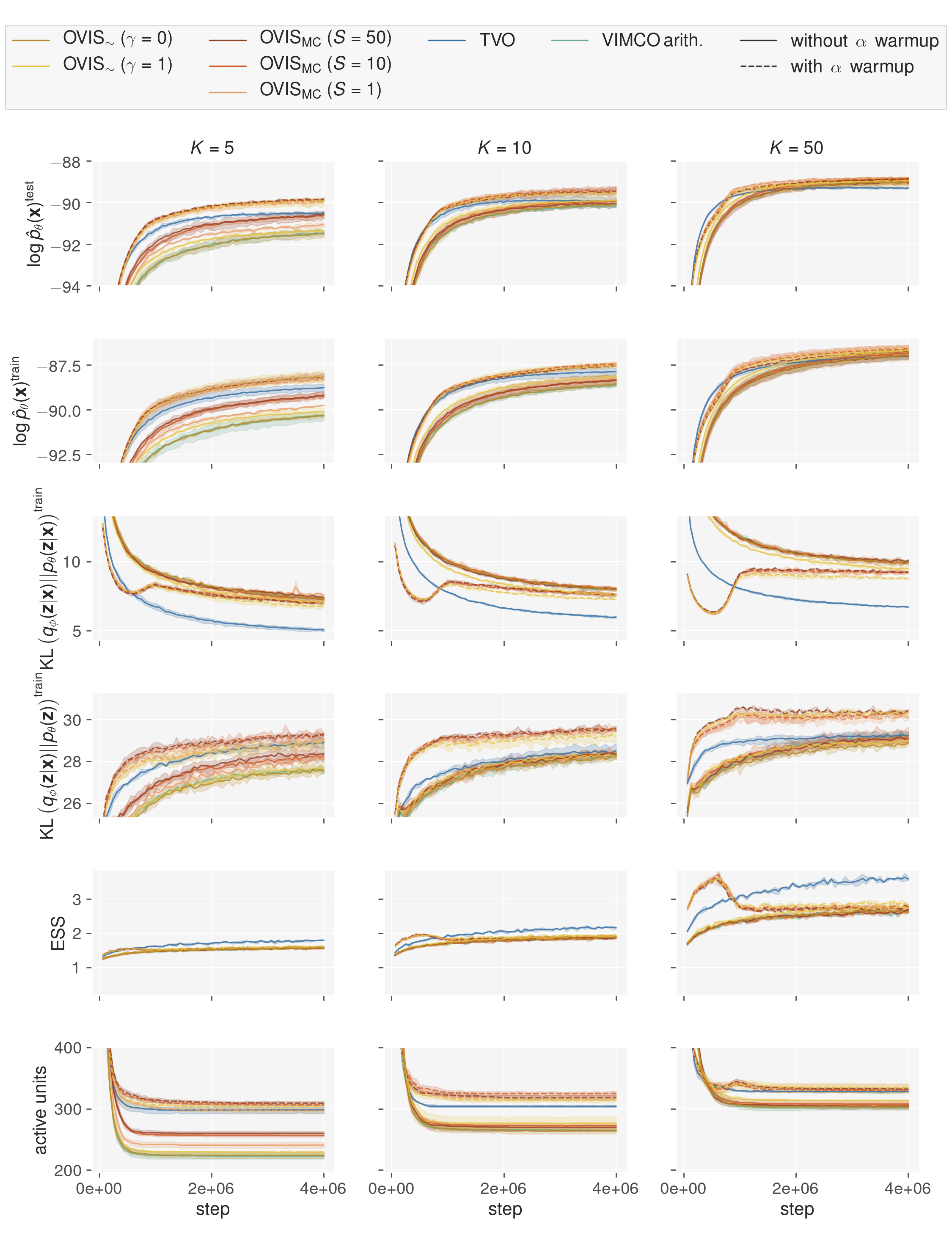}
  \captionof{figure}{Training curves for the Sigmoid Belief Network using $K = [5, 10, 50]$ particles, using two initial random seeds, with and without using the IWR bound. The number of active units is evaluated as $\mathrm{AU}=\sum_{d=1}^{D} \mathbbm{1}\left\{\operatorname{Cov}_{p(\mathbf{x})}\left(\mathbb{E}_{q_{\phi}(\boldsymbol{z} | \mathbf{x})}\left[\boldsymbol{z}_{d}\right]\right) \geq 0.01\right\}$~\citep{Burda2015-wt} using 1000 MC samples for each element of a randomly sampled subset of 1000 data points. Warming up the model by optimizing for the IWR bound with $\alpha > 0$ allows activating a larger number of units and results in models scoring higher training likelihoods.}
\end{center}

\newpage 

\subsection{Gaussian Variational Autoencoder}\label{apdx:gaussian-vae}

\begin{center}
  \captionsetup{type=figure}
  \includegraphics[width=\textwidth]{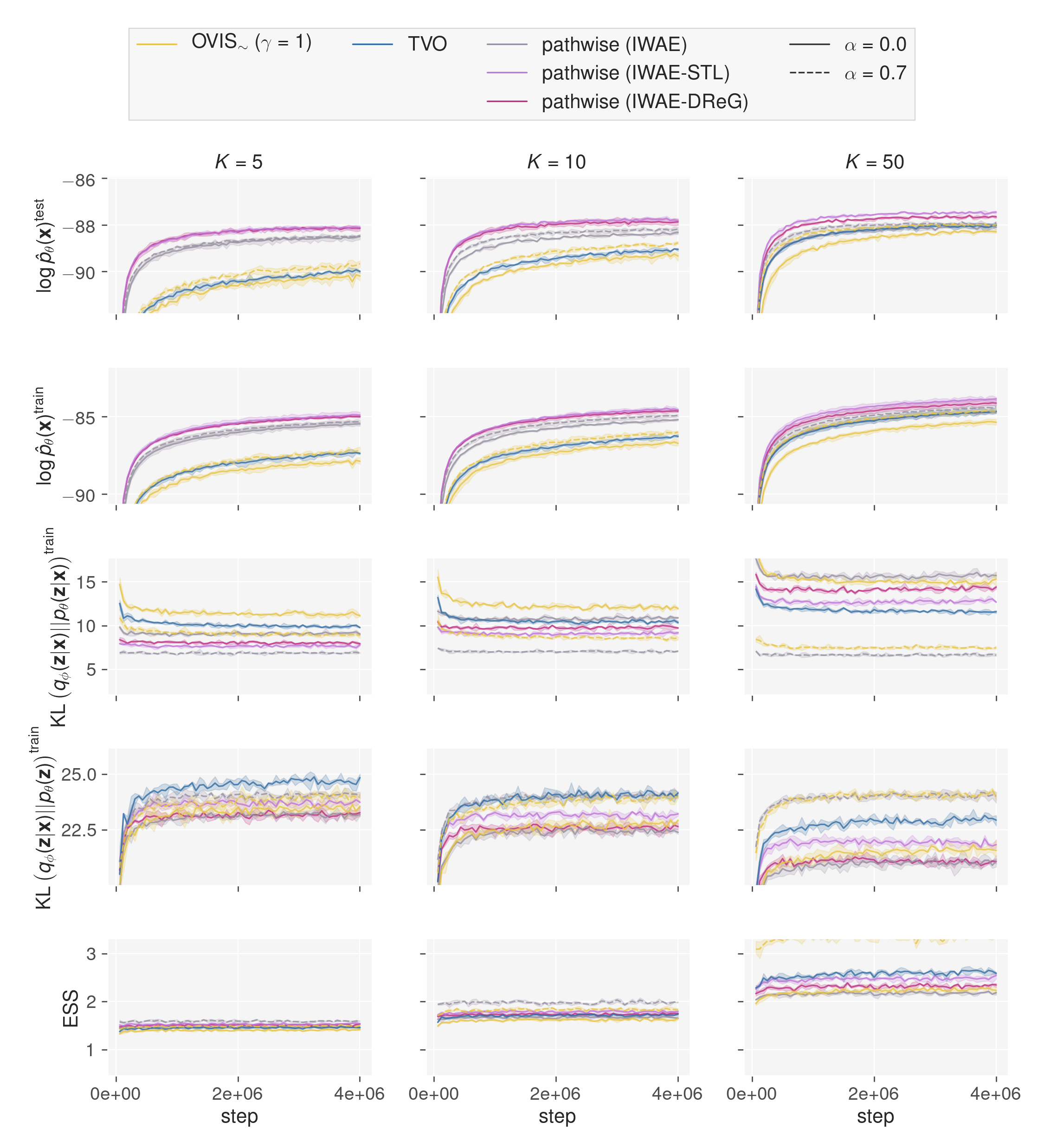}
  \captionof{figure}{Training curves a Gaussian VAE using $K = [5, 10, 50]$ particles and using two initial random seeds. The $\ovis$ estimators are used in tandem with the IWR bound with $\alpha$ fixed to $0.3$. $\ovis$ for the IWR bound yields high-quality inference networks, as measured by the divergence $\kl \left( \pth(\zb | \xb) || \qphi(\zb | \xb) \right)$.
  }
\end{center}

\newpage
\section{Implementation Details for $\ovisinf$}\label{apdx:implementation}

In order to save computational resources for large $K$ values, we implement the following factorization
\begin{equation}
    \log \ZK - \log \ZKnokunb = \log \frac{1 - 1/K}{1 - \vk} \ .
\end{equation}

In order to guarantee computational stability, we clip the normalized importance weights $\vk$ using the default PyTorch value $\epsilon = 1.19e^{-7}$. The resulting gradient estimate, used in the main experiments, is
\begin{equation}
   \grad := \sum_k \left( \log \frac{1 - 1/K}{1 - \min(1-\epsilon, \vk) } + (\gamma-1)  \vk - (1 - \gamma) \log(1 - 1/K) \right) \hk \ .
\end{equation}

\begin{figure}[h]
        \centering
        \includegraphics[width=0.8\linewidth]{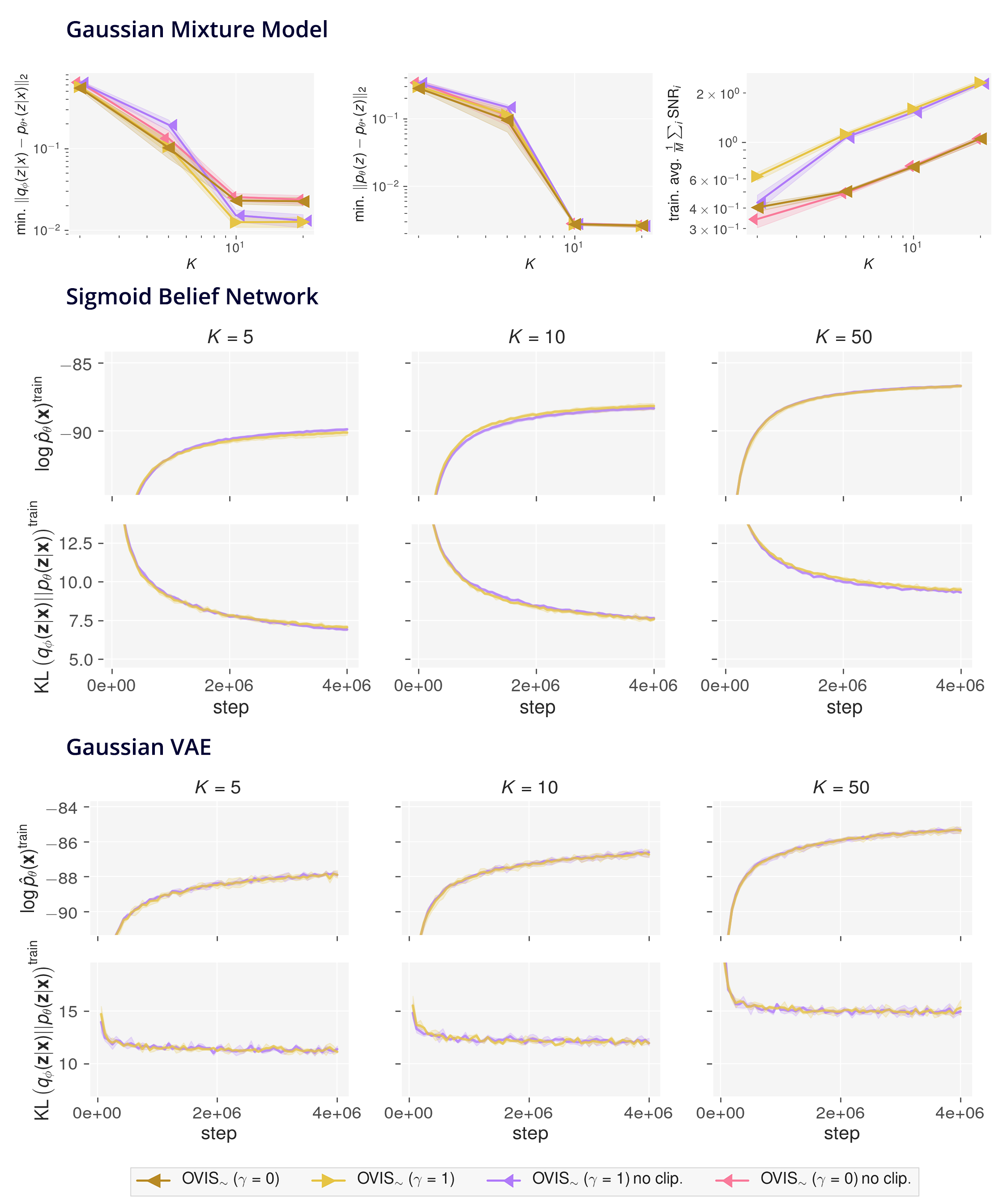}
  \caption{Effect of the importance weight clipping. Training the Gaussian Mixture Model, Sigmoid Belief Network and Gaussian VAE with and without clipping.}\label{fig:no-clip}
\end{figure}

Clipping the normalized importance weights can be interpreted as an instance of truncated importance sampling. Hence, the value of $\epsilon$ must be carefully selected. In the figure \ref{fig:no-clip}, we present a comparison of $\ovisinf$ with and without clipping. The experiments indicate that the difference is insignificant when using the default $\epsilon$.

\end{document}